\documentclass[11pt,a4paper]{article}
\usepackage{times,latexsym}
\usepackage{url}

\usepackage{graphicx}
\usepackage{amsmath,amsfonts,amssymb,amstext,bm}
\usepackage{bbm}
\usepackage{booktabs,multirow}

\usepackage[acceptedWithA]{tacl2021v1}
\usepackage{inconsolata}

\usepackage{mathrsfs}
\usepackage{paralist}
\usepackage{booktabs}
\usepackage{multirow}
\usepackage{pgfplots}
\usepackage{scalefnt}
\usepackage{colortbl}
\usepackage{xcolor}
\usepackage{array}
\usepackage{stfloats}
\usepackage{siunitx}
\usepackage{graphicx}
\usepackage{makecell}
\usepackage[ruled]{algorithm2e}
\usepackage{xspace,mfirstuc,tabulary}

\newif\iftaclinstructions
\taclinstructionsfalse 
\iftaclinstructions
\renewcommand{\confidential}{}
\renewcommand{\anonsubtext}{(No author info supplied here, for consistency with
TACL-submission anonymization requirements)}
\newcommand{\instr}
\fi

\iftaclpubformat 

\else

\fi

\definecolor{olivegreen}{rgb}{0.2,0.5,0.2}
\title{Retrieval-style In-Context Learning for Few-shot Hierarchical Text Classification}

\author{
  Huiyao Chen$^{1,}$\Thanks{Equal contribution}~~,
  Yu Zhao$^{2,*}$,
  Zulong Chen,
  Mengjia Wang,
  \\
  \textbf{ Liangyue Li,
  Meishan Zhang$^{1,}$\Thanks{The corresponding author}~~,
  Min Zhang$^1$}
  \\
  \ \\
  $^1$Institute of Computing and Intelligence, Harbin Institute of Technology (Shenzhen), China
  \\
  $^2$College of Intelligence and Computing, Tianjin University, China
  \\
  \texttt{chenhy1018@gmail.com, zhaoyucs@tju.edu.cn, chenzulong198867@163.com}
  \\
  \texttt{mason.zms@gmail.com, zhangmin2021@hit.edu.cn}
}

\date{}

\begin{document}
\maketitle
\begin{abstract}

Hierarchical text classification (HTC) is an important task with broad applications, while few-shot HTC has gained increasing interest recently.
While in-context learning (ICL) with large language models (LLMs) has achieved significant success in few-shot learning, it is not as effective for HTC because of the expansive hierarchical label sets and extremely-ambiguous labels.
In this work, we introduce the first ICL-based framework with LLM for few-shot HTC.
We exploit a retrieval database to identify relevant demonstrations, and an iterative policy to manage multi-layer hierarchical labels.
Particularly, we equip the retrieval database with HTC label-aware representations for the input texts, which is achieved by continual training on a pretrained language model with masked language modeling (MLM), layer-wise classification (CLS, specifically for HTC), and a novel divergent contrastive learning (DCL, mainly for adjacent semantically-similar labels) objective.
Experimental results on three benchmark datasets demonstrate superior performance of our method, and we can achieve state-of-the-art results in few-shot HTC.

\end{abstract}

\section{Introduction}

Hierarchical text classification (HTC), a specialized branch of multilabel text classification, involves the systematic arrangement and categorization of textual data throughout a tiered label structure.
The output labels are organized in a parent-child hierarchy, with the higher-level labels encompassing broader concepts, and the child labels delineating more specific subtopics or attributes. 
In recent years, HTC has gained significant attention, due to its applicability across a variety of fields, including recommendation systems \cite{DBLP:journals/artmed/SunNYGZ23, DBLP:conf/www/AgrawalGPV13}, document categorization \cite{DBLP:journals/bioinformatics/PengYWZMZ16,DBLP:conf/icmla/KowsariBHMGB17} and information retrieval \cite{sinha-etal-2018-hierarchical}.

In standard supervised HTC, there is an underlying assumption of abundant training samples \cite{DBLP:journals/taslp/ZhaoWHD23, DBLP:conf/aaai/ImKOJK23, DBLP:conf/acl/SongWY23}, which is often unattainable and expensive to construct manually.
Moreover, HTC datasets are characterized by a complex hierarchical label structure, with leaf labels typically following a Zipfian distribution, resulting in very few data instances for these labels.
As a result, the few-shot setting is more realistic, and has gained increasing interest recently \cite{ji-etal-2023-hierarchical,DBLP:conf/acl/BhambhoriaCZ23, DBLP:conf/acl/WangQLCZLZZ23}.
Nevertheless, existing works often struggle with unsatisfactory performance in this setting.
For example, BERT with the vanilla fine-tuning strategy performs extremely poorly in few-shot HTC.

\begin{figure}[t]
    \centering
    \includegraphics[width=1.00\linewidth]{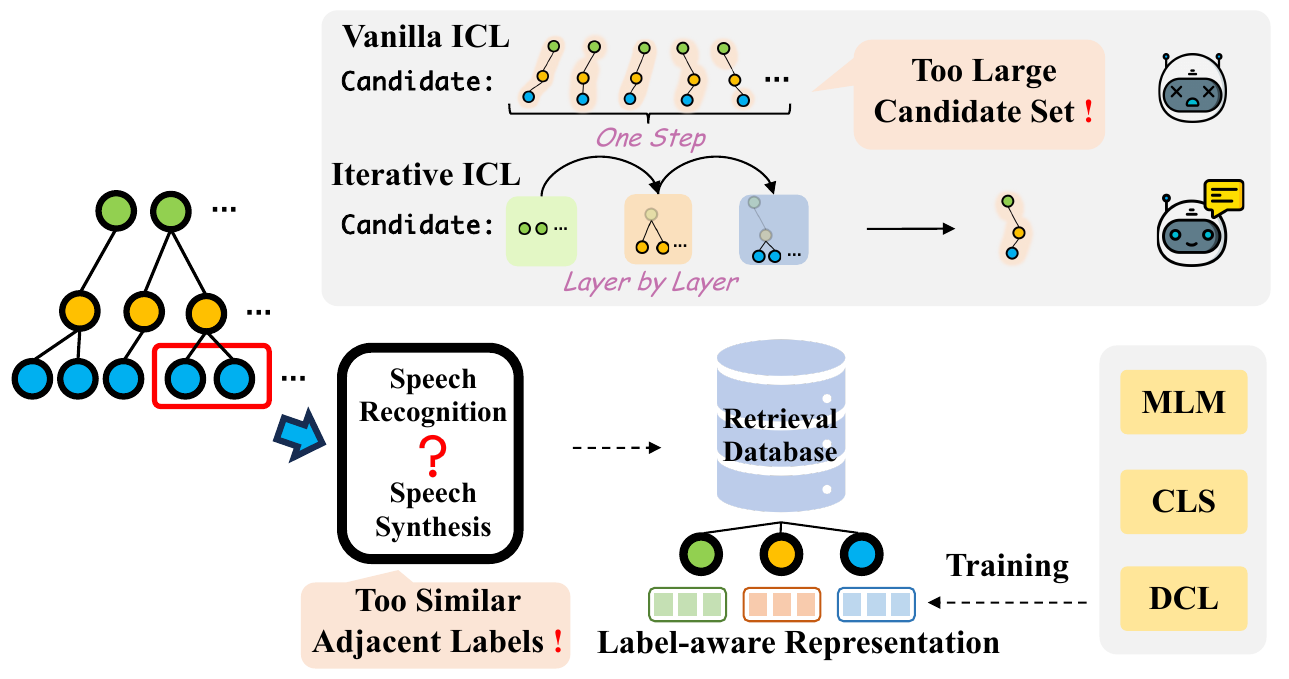}
    \caption{The problems of ICL-based few-shot HTC and our solutions. MLM, CLS and DCL denote \textbf{M}ask \textbf{L}anguage \textbf{M}odeling, Layer-wise \textbf{CL}a\textbf{S}sification and \textbf{D}ivergent \textbf{C}ontrastive \textbf{L}earning, which are the three objectives for indexer training.}
    \label{fig:intro}
\end{figure}

Recently, large language models (LLMs) have achieved notable success on various NLP tasks \cite{DBLP:journals/corr/abs-2304-10428, DBLP:conf/iclr/DrozdovSASSCBZ23, DBLP:journals/corr/abs-2307-04408},
which have significantly enhanced the efficacy of in-context learning (ICL) with relevant demonstrations in the few-shot setting \cite{DBLP:journals/corr/abs-2305-14622,DBLP:conf/acl/DaiS0HMSW23,DBLP:conf/eacl/ZhangWYYVL23}.
However, the application of ICL on HTC faces unique challenges, diverging from traditional text classification scenarios.
These challenges are primarily due to two distinct characteristics of HTC, as delineated in Figure \ref{fig:intro}.
Firstly, HTC features a deep hierarchical labeling structure and expansive label sets, resulting in large label sets in ICL, which adversely impacts its performance.
Secondly, as the hierarchy deepens, the semantic similarity between adjacent labels increases \cite{DBLP:journals/isci/SteinJV19}, making it very challenging to select relevant demonstrations that guide the learning process efficiently.

In this work, we introduce the first ICL-based framework for few-shot HTC.
Specifically, we use a LLM as the foundation model for inference, and provide demonstrations to guide HTC label generation through ICL.
Our success depends on finding suitable demonstrations for a given input.
In order to achieve this, we build a retrieval database that can find the most-relevant demonstrations for the input.
Further, in order to avoid providing an enormous set of multi-layer contextual HTC labels all at once, as is required for ICL,
we suggest an iterative policy to infer the labels layer-by-layer, reducing the number of candidate labels greatly.



The quality of our retrieval database is highly critical.
The key idea is to obtain the HTC label-aware representations for input texts,
which are then used for subsequent retrieving.
Given an input, we define prompt templates for each-layer HTC, concatenating them with the raw input to form a new text.
The hidden vectors of prompts are exploited as label-aware representations.
We perform continual training to learn our representations using three types of objectives: the masked language model (MLM),
the multi-label classification (MLC) for HTC particularly and a well-designed divergent contrastive learning (DCL) objective.
The DCL is especially useful for the semantically-closed HTC labels (e.g., adjacent labels from the same parent).
In addition, we can incorporate label descriptions naturally by DCL,
which can benefit ICL-based HTC much.

We conduct experiments on two classic English hierarchical classification datasets, WOS \cite{wos} and DBpedia \cite{sinha-etal-2018-hierarchical},
as well as a Chinese patent dataset\footnote{This dataset is not allowed to be publicly released due to local law.}, and measure model performance by both micro-F1 and macro-F1 metrics consistent with previous work.
Results show that our method is highly effective, giving improved results compared with a series of baselines for the few-shot HTC with different shots.
We can also achieve state-of-the-art (SOTA) results on these HTC datasets.
Further, we perform thorough qualitative analysis in order to gain a comprehensive understanding of our methodology.

In summary, our contributions can be summarized as follows:
\begin{compactitem}
    \item We present the first ICL-based work for HTC backended with LLMs, utilizing the knowledge from the existing demonstrations during the LLM inference.
    \item We propose a novel retrieve-style framework for ICL of HTC, which can help to find highly-relevant demonstrations to support HTC label prediction.
    \item We conduct extensive experiments to compare our method with several representative baselines, and results show that our method can advance the SOTA of HTC further.
\end{compactitem}

Our code is publicly available at \href{https://github.com/DreamH1gh/TACL2024}{https://github.com/DreamH1gh/TACL2024} to facilitate future research.

\section{Related Work}
HTC is initially presented by \cite{DBLP:conf/icml/KollerS97}, and neural network models have achieved great advances in this task.
Previous approaches treat HTC as multi-label classification and adopt traditional text classification methods for HTC problems \cite{DBLP:conf/acl/AlyRB19, DBLP:journals/eswa/ZhangXSC22}. 
The majority of recent HTC studies pay the focus on finding ways to insert the hierarchical label knowledge into the model \cite{chen-etal-2021-hierarchy,DBLP:conf/emnlp/MaoTHR19,sinha-etal-2018-hierarchical}.
Several works attempt to solve HTC problems by modeling the hierarchical labels as a graph or a tree structure \cite{DBLP:conf/acl/ZhouMLXDZXL20, DBLP:conf/acl/TaiSM15}
while other researchers try to apply meta-learning \cite{DBLP:conf/emnlp/WuXW19} or reinforcement learning \cite{DBLP:conf/emnlp/MaoTHR19} to leverage HTC label structure.

However, existing methods mainly concentrate on encoding the holistic label structure, ignoring the classification of nuanced long-tail terminal labels.
There have been efforts prove that retrieval augmented methods could help classification task with only few-shot samples \cite{chen2022contrastnet,zhang-etal-2022-prompt-based,DBLP:conf/acl/YuZZ0SZ23,DBLP:journals/isci/XiongYNL23}, which could be a solution of long-tail challenge.
Given this insight, we make an attempt to convert the HTC task to a retrieval form.

Moreover, with the development of LLMs, recent work explores solutions that tackle traditional NLP tasks with the ICL paradigm and achieve surprising effectiveness \cite{DBLP:journals/corr/abs-2305-14622, DBLP:conf/acl/FeiHCB23,DBLP:conf/emnlp/MinLHALHZ22,DBLP:conf/acl-deelio/LiuSZDCC22}.
But ICL strongly relies on the demonstration selecting \cite{DBLP:conf/acl/GaoFC20,DBLP:conf/icml/ZhaoWFK021,rubin-etal-2022-learning,zhang-etal-2022-active,li-etal-2023-unified}.
Many studies explore adjusting demonstrations for better performance through instruction formatting \cite{DBLP:conf/iclr/ZhouMHPPCB23}, examples ordering \cite{liu2021makes} and demonstration filtering \cite{DBLP:conf/acl/SorensenRRSRDKF22}.
In our work, we combine the ICL-based framework with retrieval for HTC, selecting demonstrations that involve both the language knowledge of LLM and advantages of retrieval.
\section{Method}

\begin{figure*}[t]
    \centering
    \includegraphics[width=1.00\linewidth]{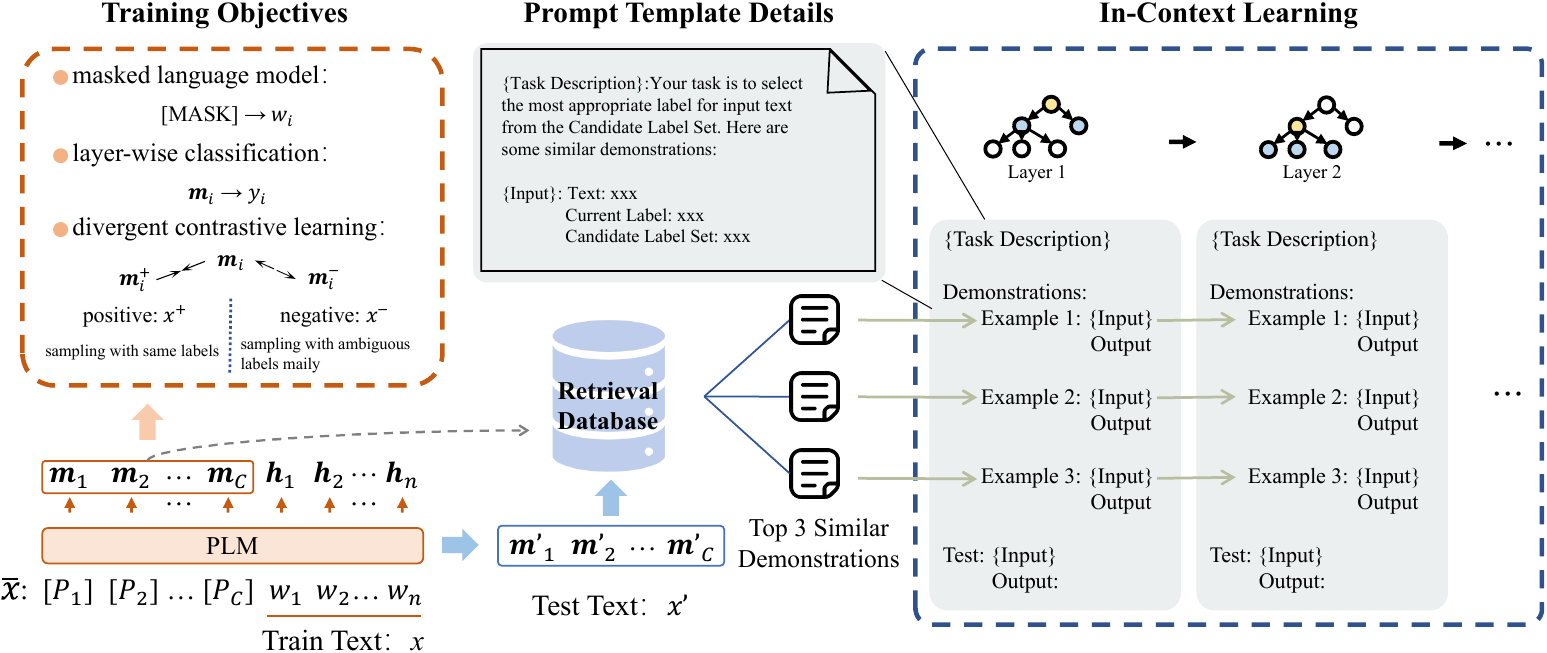}
    \caption{The architecture of retrieval-style in-context learning for HTC. The \texttt{[P$_j$]} term is a soft prompt template token to learn the $j$-th hierarchical layer label index representation.}
    \label{fig:mdoel}
\end{figure*}

\paragraph{Problem Formulation}
In HTC tasks, the structure of labels $\mathcal{H} = (\mathcal{Y},E)$ is often predefined as a tree,
where $\mathcal{Y}=\{Y_1, Y_2, \dots, Y_L\}$ is a set of nodes (labels) and $E$ indicates the parent-child hierarchical edges (connections) between the labels.
It is worth noting that in the label structure, every node, except for the root, has one and only one parent.
Generally speaking, HTC tasks select the label path in $\mathcal{H}$ for a given text $x$.
We define that $x=w_1w_2 \cdots w_n$ is a text and $y=\{y_1,y_2,\dots,y_C\} \subseteq \mathcal{Y}$ is the corresponding hierarchical labels which follow $\mathcal{H}$, where
$C$ denotes the maximum label depth.

\paragraph{Proposed Framework}

Figure \ref{fig:mdoel} illustrates our ICL-based framework for HTC.
We first train a PLM-based indexer and build a retrieval database containing reference samples (the training data).
After that, we perform a similarity search in the retrieval database with the text to be inferred.
Finally, we construct an ICL prompt with highly similar demonstrations for prediction.

We will introduce our ICL prompt policy for HTC ($\S ~\ref{sec:in_context_learning}$), and then detail the retrieval database construction ($\S ~\ref{sec:retri_database}$) and demonstration retrieval methods ($\S ~\ref{sec:demons_retri}$).

\subsection{In-Context Learning}
\label{sec:in_context_learning}

In order to integrate label structural information into ICL, we propose an iterative approach by decoupling the label structure $\mathcal{H}$.
We decompose the label structure into several subclusters, each corresponding to a parent-child set.
Then, we employ an iterative method to produce the sub-labels layer by layer until we arrive at the leaf labels.

As shown in Figure \ref{fig:mdoel}, we perform iterative inference at each hierarchy level.
Based on the Top K similar demonstrations, the prompt contains K identical structured text blocks.
Each block contains three parts: \texttt{Text}, \texttt{Current Label}, and \texttt{Candidate Label Set}.
\texttt{Text} is the demonstration content.
\texttt{Current Label} is the predicted label of the previous hierarchy level\footnote{\texttt{Current Label} is Root when predicting the first hierarchy level}.

When the LLM is used for inference in classification tasks, the entire set of labels is always presented.
The inference result is drawn from this large label set.
In contrast, our method supplies a pruned subset of labels as a concise candidate label set.
\texttt{Candidate Label Set} is the intersection of the child nodes of the current label and the selected K demonstration labels, which maximizes the use of demonstration information and avoids the impact of erroneous labels.
The predicted label of the next hierarchy level is required to be selected from the candidate label set.

\subsection{Retrieval Database Construction}
\label{sec:retri_database}

After determining the ICL prompt policy, it is crucial to obtain demonstrations related to the test text, which will provide effective guidance for LLM inferences.
Firstly, we train a HTC indexer to generate index vectors for each training sample.
We employ a pretrained text encoder as the indexer and use a prompt template to elicit multi-layer representations as index vectors.
To make the index vectors discriminative, the indexer is trained via DCL based on label descriptions.

\paragraph{Index Vector Representation.}

To further utilize the language knowledge embedded in pre-trained text encoders and leverage interdependencies among hierarchical labels, we propose the construction of a concise prompt template prior to raw input $x$.
The new text is formatted as:
$\bar{x}$=\texttt{[P$_1$] [P$_2$] \dots [P$_C$]} $x$.
Here, the \texttt{[P$_j$]} term is a soft prompt template token to learn the $j$-th hierarchical layer label index representation.
Then, we input $\bar{x}$ into the encoder of PLM to obtain the hidden states:
\begin{equation}
\label{eq:1}
    \bm{m}_1 \cdots \bm{m}_C~\bm{h}_1 \cdots \bm{h}_n = \text{PLM}(\bar{x}).
\end{equation}
Thus, we can obtain the index vectors $\bm{m}_1 \cdots \bm{m}_C$ consisting of hidden state embeddings for all fixed-position \texttt{[P]},
where we consider $\bm{m}_j$ as the index vector of the $j$-th hierarchical level corresponding to $x$.

\begin{figure}[t]
    \centering
    \includegraphics[width=1.00\linewidth]{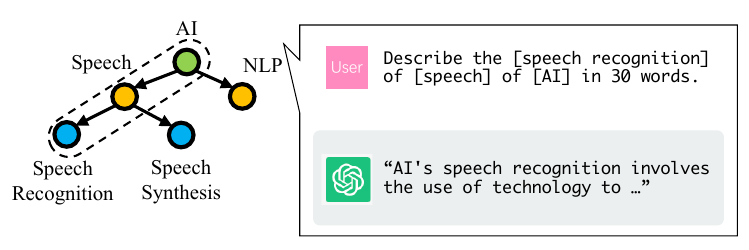}
    \caption{Label description generation.}
    \label{fig:label}
\end{figure}

\paragraph{Label Description.}
In order to reduce the ambiguity errors caused by insufficient label information,
we explore diverse approaches that aim to provide more informative and representative label information for HTC task.
First, we propagate the textual information of all label nodes to their corresponding leaf nodes,
obtaining the textual information with the entire label path.
As shown in Figure \ref{fig:label}, for the original leaf label ``speech recognition'', its label path is ``speech recognition of speech of AI''.

However, due to the close semantic proximity of adjacent leaf node labels, the generated label path may still be insufficient or ambiguous.
For example, ``speech recognition of speech of AI'' and ``speech synthesis of speech of AI'' may still be difficult to distinguish.
To address this issue, we use the LLM to expand and enhance the label path $l$ of $x$ by leveraging the knowledge contained within the LLM:
\begin{equation}
    \label{eq:descirbe}
    d = \text{LLM}(\texttt{Describe},l),
\end{equation}
where $d$ is the description of the label path $l$ and \texttt{Describe} denotes the prompt used to generate the description.
By utilizing expanded and enhanced label descriptions, we could obtain a more detailed explanation of the label.

\paragraph{Indexer Training.}

For indexer training, we apply the objectives of mask language modeling $\mathcal{L}_{mlm}$, and layer-wise classification $\mathcal{L}_{cls}$.
$\mathcal{L}_{mlm}$ is used to predict the words that fill the random mask tokens in the inputs.
$\mathcal{L}_{cls}$ is to predict HTC labels through each hierarchical layer index vectors.

Additionally, we propose DCL for indexer training, which uses label text information to select positive and negative samples.
For $x$, positive samples are chosen from sentences with the same label as $x$.
Additionally, the corresponding label description $d$ could be treated as a positive sample.
Negative samples consist of two parts.
First, based on the similarity between $d$ and descriptions of other labels,
negative examples are sampled from highly similar label categories.
Similarly, their corresponding label descriptions could be also treated as negative samples.
In addition, a few randomly selected sentences from other labels are used as negative samples of $x$.
Thus, compared to traditional random sampling methods,
our negative sample selection approach opts for more instances with semantically similar labels as hard negative samples.

Then the index vectors among the positive samples are pulled together and the negative ones are pushed apart.
Taking $x$ as an example, denote $\bm{B}=\{x, x^{+}, x^{-}_{1}, \dots, x^{-}_{n}\}$ as a group of input data.
The contrastive loss can be calculated as:
\begin{equation}
    \label{eq:cl}
    \mathcal{L}_{\text{con}}=-\sum_{j}^{C}\log \frac{e^{\cos(\bm{m}_j,\bm{m}_{j}^+)/\tau}}{\sum_{k}^{n}e^{\cos(\bm{m}_j,\bm{m}_{j,k}^-)/\tau}},
\end{equation}
where $cos(\cdot ,\cdot)$ is the cosine similarity, $\tau$ is the contrastive learning temperature.
In comparison to calculating the contrastive loss in a random sampling batch, our DCL pays more attention to samples whose labels are less similar to the $x$.

The final objective is set in the multi-task form:
\begin{equation}
    \label{eq:loss}
    \mathcal{L} = \mathcal{L}_{\text{mlm}} + \alpha\mathcal{L}_{\text{cls}} + \beta \mathcal{L}_{\text{con}}.
\end{equation}

After the training step, we store index vectors $\bm{m_1} \cdots \bm{m_C}$ of each training instance to construct the retrieval database.

\subsection{Demonstration Retrieval}
\label{sec:demons_retri}

With the database and indexer in hand, we can process predictions in retrieval form.
For the test text $x'$, we also use the trained indexer to obtain the hierarchical index vectors $\bm{m}'_{1} \cdots \bm{m}'_{C}$.
Then, we select similar instances from the retrieval database by calculating similarity between their index vectors.
For each training instance $x$, we have $C$ index vectors $\bm{m}_1 \cdots \bm{m}_C$ in retrieval database.
The similarity between $x$ and $x'$ can be calculated as:
 \begin{equation}
    \operatorname{sim}(x,x') = \sum^{C}_{j}\frac{2^{j-1}}{2^C-1} \cdot \cos(\bm{m}_j, \bm{m}'_{j}),
\end{equation}
where the first factor is utilized to adjust the weights of similarity between different hierarchical layer, while ensuring that $\sum^{C}_{j}\frac{2^{j-1}}{2^C-1}=1$.
As the hierarchy deepens, the impact of index vector similarity gradually increases.
Then, we choose the Top K most similar instances from the database as demonstrations.
It is worth noting that we filter out instances with the same label here, to ensure that the labels of the Top K instances are different, providing relatively diverse instances for ICL.

\section{Experiments}

\subsection{Settings}

\setlength{\tabcolsep}{5pt}
\begin{table}[t]
\begin{center}
\resizebox{1\columnwidth}{!}{
\begin{tabular}{lccc}
\toprule
\multirow{2}{*}{\textbf{Statistics}} & \multirow{2}{*}{\textbf{WOS}} & \multirow{2}{*}{\textbf{DBpedia}} & \multirow{2}{*}{\textbf{Patent}} \\
& & &  \\
\midrule
\#levels                  & 2      &       3 & 4     \\
\#Number of documents     & 46,985 & 381,025 & 30,104\\ \midrule
\#Level 1 Categories      & 7      & 9       & 10    \\
\#Level 2 Categories      & 134    & 70      & 17    \\
\#Level 3 Categories      & NA     & 219     & 105   \\
\#Level 4 Categories      & NA     & NA      & 305   \\ \midrule
\#Mean label length       & 1.8    & 1.7     & 4.4   \\
\#Max label length        & 3      & 7       & 14    \\
\#Mean document length    & 200.7  & 106.9   & 335.1 \\
\#Max document length     & 1262   & 881     & 1669  \\

\bottomrule
\end{tabular}
}
\caption{Overview of HTC datasets.}
\label{tab:data}
\end{center}
\end{table}

\paragraph{Dataset and Evaluation Metrics.}
Our experiments are evaluated on three datasets:
Web-of-Science (WOS) \cite{wos},
DBpedia \cite{sinha-etal-2018-hierarchical}
and Patent.
WOS and DBpedia are both widely used English datasets for HTC and Patent which we collected consists of 30,104 Chinese patent records.
We evaluate the effectiveness of our proposed method on both English and Chinese datasets.
All of them are for single-path HTC.
The statistics are illustrated in Table \ref{tab:data}.
Following the previous work, we report experimental results with Micro-F1 and Macro-F1.

\paragraph{Model Details.}
We utilize bert-base-uncased\footnote{\href{https://huggingface.co/bert-base-uncased}{https://huggingface.co/bert-base-uncased}} \cite{DBLP:conf/naacl/DevlinCLT19} as the base indexer for WOS and DBpedia datasets, while for Patent dataset, we employ chinese-bert-wwm-ext\footnote{\href{https://huggingface.co/hfl/chinese-bert-wwm-ext}{https://huggingface.co/hfl/chinese-bert-wwm-ext}} \cite{chinese-bert-wwm, cui-etal-2020-revisiting}.
Regarding LLM, we select vicuna-7b-v1.5-16k\footnote{\href{https://github.com/lm-sys/FastChat}{https://github.com/lm-sys/FastChat}} \cite{zheng2023judging} and gpt-3.5-turbo-0613\footnote{\href{https://openai.com/blog/chatgpt}{https://openai.com/blog/chatgpt}. ICL inference in experiments is based on this model unless otherwise specified.},
which performs well on English for WOS and DBpedia datasets,
and ChatGLM-6B\footnote{\href{https://github.com/THUDM/ChatGLM-6B}{https://github.com/THUDM/ChatGLM-6B}} \cite{zeng2022glm, du2022glm}, the top-performing open-source Chinese language model for Patent (due to legal restrictions, we can only evaluate it on open-source models).
Our model is implemented with the OpenPrompt toolkit \cite{ding-etal-2022-openprompt}.

\paragraph{Experimental Settings.}
As mentioned in the introduction, we try to validate the effectiveness of our proposed method on the few-shot classes in the long-tail phenomenon.
Specifically, we focus on the few-shot setting, where only Q samples per label path are available for training and use the same seeds as \citet{ji-etal-2023-hierarchical}, as shown in Algorithm \ref{algorithm:sample}.
We conduct experiments based on $\text{Q} \in \{1,2,4,8,16\}$.
The batch size of our proposed method is 1.
It is composed of a training sample, a randomly selected positive sample from the same label, 4 randomly selected negative samples from the Top4 labels based on label description similarity, and 10 randomly selected negative samples from other labels.
For all datasets, the learning rate is $5*10^{-5}$ and we train the model for 20 epochs and apply the Adam optimizer \cite{kingma2014adam} with a linearly decaying schedule with warmup steps at 0.
The temperature of Vicuna, GPT3.5 and ChatGLM are both 0.2.
$\alpha$ in Equation \ref{eq:loss} is 1 and $\beta$ is 0.01.

\begin{algorithm}[t]
    \small
    \caption{Sampling for HTC Few-shot}
    \label{algorithm:sample}
    \LinesNumbered
    
    \KwIn{
        Shot number: Q, Complete HTC dataset: $\mathcal{D}={\{(x_i,y_i)\}}^N_{i=1}$
    }
    \KwOut{
        Q-shot sampling dataset: $\mathcal{S}$
    }
    // Categorize samples by label path\\
    Label path dictionary: $\mathcal{C} = \{\}$\;
    \For{i = 1 to N}{
        \eIf{$y_i$ not key in $\mathcal{C}$}{
            $\mathcal{C}$ = $\mathcal{C} \cup \{y_i: \{x_i\}\}$\;
        }{
            $\mathcal{C}[y_i]$ = $\mathcal{C}[y_i] \cup \{x_i\}$\;
        }
    }
    // Q-shot random sampling \\
    Q-shot sampling dataset: $\mathcal{S} = \{\}$\;
    \For{i = 1 \dots until all keys in $\mathcal{C}$ are traversed}{
        \eIf{$\text{Count}(\mathcal{C}[y_i]) \leq Q$}{
            $\mathcal{S}$ = $\mathcal{S} \cup \mathcal{C}[y_i]$\;
        }{
            
            $\mathcal{S}$ = $\mathcal{S} \cup \text{Random Sample}(\mathcal{C}[y_i], Q)$\;
        }
    }
    \Return $\mathcal{S}$
\end{algorithm}

\paragraph{Baselines.}
In this work, we select several recent works as baselines:

\begin{itemize}
    \item \textbf{BERT} with vanilla fine-tuning transforms HTC into a multi-label classification task. It is a standard method for HTC.

    \item \textbf{HiMatch} \cite{chen-etal-2021-hierarchy} learns the representation of text and labels separately and then defines different optimization objectives based on them to improve HTC.

    \item \textbf{HGCLR} \cite{wang-etal-2022-incorporating} incorporates the hierarchical label structure directly into the text encoder and obtains the hierarchy-aware text representation for HTC.

    \item \textbf{HPT} \cite{wang-etal-2022-hpt} leverages a dynamic virtual template with soft-prompt label words and a zero-bounded multi-label cross-entropy loss, ingeniously aligning the goals of HTC and MLM.

    \item \textbf{HierVerb} \cite{ji-etal-2023-hierarchical} treats HTC as a multi-label problem at different levels, utilizing vectors as constrained by the hierarchical structure, effectively integrating knowledge of hierarchical labels.

    \item \textbf{EPR} \cite{rubin-etal-2022-learning} estimates the output probability based on the input and a candidate training example prompt, separating examples as positive and negative and allowing effective retrieval of training examples as prompts during testing.

    \item \textbf{REGEN} \cite{DBLP:conf/acl/YuZZ0SZ23} employs a retrieval model and a classification model, utilizing class-specific verbalizers and a general unlabeled corpus to enhance semantic understanding. Notably, REGEN incorporates supplementary unsupervised data\footnote{We employ full dataset texts, excluding the training data, as unsupervised data.}.
\end{itemize}

\begin{table*}[t]
    \centering
    \fontsize{8}{8}\selectfont
    \setlength{\tabcolsep}{1.mm}
        \begin{tabular}{c l c c c c}
            \toprule
            
            \multirow{2}{*}{Q} & \makecell[c]{\multirow{2}{*}{Method}} & \multicolumn{2}{c}{\textbf{WOS(Depth 2)}} & \multicolumn{2}{c}{\textbf{DBpedia(Depth 3)}} \\ \cmidrule{3-6}
                               &                                       &       Micro-F1      &       Macro-F1      &       Micro-F1       &        Macro-F1     \\ \midrule
            
            \multirow{9}{*}{1} &              BERT                               ${\dagger}$   &           2.99   $\pm$ 20.85{\color{olivegreen}(5.12)}  &           0.16  $\pm$ 0.10 {\color{olivegreen}(0.24)}  &           14.43  $\pm$ 13.34 {\color{olivegreen}(24.27)} &           0.29   $\pm$ 0.01 {\color{olivegreen}(0.32)}   \\
                               &  HiMatch                                        ${\dagger}$   &           43.44  $\pm$ 8.09 {\color{olivegreen}(48.26)} &           7.71  $\pm$ 4.90 {\color{olivegreen}(9.32)}  &                                 -                        &                               -                          \\
                               &  HGCLR                                          ${\dagger}$   &           9.77   $\pm$ 11.77{\color{olivegreen}(16.32)} &           0.59  $\pm$ 0.10 {\color{olivegreen}(0.63)}  &           15.73  $\pm$ 31.07 {\color{olivegreen}(25.13)} &           0.28   $\pm$ 0.10 {\color{olivegreen}(0.31)}   \\
                               &  HPT                                            ${\dagger}$   &           50.05  $\pm$ 6.80 {\color{olivegreen}(50.96)} &           25.69 $\pm$ 3.31 {\color{olivegreen}(27.76)} &           72.52  $\pm$ 10.20 {\color{olivegreen}(73.47)} &           31.01  $\pm$ 2.61 {\color{olivegreen}(32.50)}  \\
                               &  HierVerb                                       ${\dagger}$   &           58.95  $\pm$ 6.38 {\color{olivegreen}(61.76)} &           44.96 $\pm$ 4.86 {\color{olivegreen}(48.19)} &           91.81  $\pm$ 0.07  {\color{olivegreen}(91.95)} &           85.32  $\pm$ 0.04 {\color{olivegreen}(85.44)}  \\
                               &  EPR                                                          &           31.77  $\pm$ 3.15 {\color{olivegreen}(35.31)} &           6.61  $\pm$ 2.70 {\color{olivegreen}(9.66)}  &           16.58  $\pm$ 8.94  {\color{olivegreen}(25.60)} &           7.41   $\pm$ 4.13 {\color{olivegreen}(11.91)}  \\
                               &  REGEN                                                        &           5.62   $\pm$ 2.98 {\color{olivegreen}(8.70)}  &           2.59  $\pm$ 2.45 {\color{olivegreen}(4.71)}  &           18.70  $\pm$ 8.19  {\color{olivegreen}(27.33)} &           8.17   $\pm$ 3.87 {\color{olivegreen}(12.20)}  \\ 
                               \cmidrule{2-6}
                               &                    Retrieval                                  &\underline{63.46} $\pm$ 2.30 {\color{olivegreen}(65.99)} &\underline{50.24}$\pm$ 2.21 {\color{olivegreen}(52.66)} &\underline{93.68} $\pm$ 0.05  {\color{olivegreen}(93.74)} &\underline{88.41} $\pm$ 0.23 {\color{olivegreen}(88.67)}  \\
                               &              Retrieval-style ICL                              &\textbf{   68.91  $\pm$ 0.48 {\color{olivegreen}(69.38)}}&\textbf{   57.41 $\pm$ 0.40 {\color{olivegreen}(57.82)}}&\textbf{   94.54  $\pm$ 0.03  {\color{olivegreen}(94.58)}}&\textbf{   89.75  $\pm$ 0.09 {\color{olivegreen}(94.83)}} \\ \midrule

            \multirow{9}{*}{2} &              BERT                               ${\dagger}$   &           46.31  $\pm$ 0.65 {\color{olivegreen}(46.85)} &           5.11  $\pm$ 1.31 {\color{olivegreen}(5.51)}  &           87.02  $\pm$ 3.89  {\color{olivegreen}(88.20)} &           69.05  $\pm$ 26.81{\color{olivegreen}(73.28)}  \\
                               &  HiMatch                                        ${\dagger}$   &           46.41  $\pm$ 1.31 {\color{olivegreen}(47.23)} &           18.97 $\pm$ 0.65 {\color{olivegreen}(21.06)} &                                 -                        &                               -                          \\
                               &  HGCLR                                          ${\dagger}$   &           45.11  $\pm$ 5.02 {\color{olivegreen}(47.56)} &           5.80  $\pm$ 11.63 {\color{olivegreen}(9.63)} &           87.79  $\pm$ 0.40  {\color{olivegreen}(88.42)} &           71.46  $\pm$ 0.17 {\color{olivegreen}(71.78)}  \\
                               &  HPT                                            ${\dagger}$   &           57.45  $\pm$ 1.89 {\color{olivegreen}(58.99)} &           35.97 $\pm$ 11.89 {\color{olivegreen}(39.94)}&           90.32  $\pm$ 0.64  {\color{olivegreen}(91.11)} &           81.12  $\pm$ 1.33 {\color{olivegreen}(82.42)}  \\
                               &  HierVerb                                       ${\dagger}$   &           66.08  $\pm$ 4.19 {\color{olivegreen}(68.01)} &           54.04 $\pm$ 3.24 {\color{olivegreen}(56.69)} &           93.71  $\pm$ 0.01  {\color{olivegreen}(93.87)} &           88.96  $\pm$ 0.02 {\color{olivegreen}(89.02)}  \\
                               &  EPR                                                          &           36.04  $\pm$ 2.97 {\color{olivegreen}(39.11)} &           16.28 $\pm$ 1.94 {\color{olivegreen}(18.32)} &           21.89  $\pm$ 5.02  {\color{olivegreen}(27.02)} &           15.96  $\pm$ 2.96 {\color{olivegreen}(19.02)}  \\
                               &  REGEN                                                        &           49.55  $\pm$ 2.88 {\color{olivegreen}(52.64)} &           12.12 $\pm$ 3.54 {\color{olivegreen}(15.91)} &           87.91  $\pm$ 2.44  {\color{olivegreen}(90.57)} &           71.80  $\pm$ 2.41 {\color{olivegreen}(74.35)}  \\  
                               \cmidrule{2-6}
                               &                    Retrieval                                  &\underline{69.85} $\pm$ 0.63 {\color{olivegreen}(70.58)} &\underline{58.64} $\pm$ 0.58 {\color{olivegreen}(59.25)}&\underline{94.12} $\pm$ 0.18  {\color{olivegreen}(94.32)} &\underline{89.33} $\pm$ 0.19 {\color{olivegreen}(89.54)}  \\  
                               &              Retrieval-style ICL                              &\textbf{   71.68  $\pm$ 0.09 {\color{olivegreen}(71.76)}}&\textbf{   61.99 $\pm$ 0.10 {\color{olivegreen}(62.08)}}&\textbf{   94.87  $\pm$ 0.10  {\color{olivegreen}(94.97)}}&\textbf{   90.82  $\pm$ 0.08 {\color{olivegreen}(90.89)}} \\ \midrule
            
            \multirow{9}{*}{4} &              BERT                               ${\dagger}$   &           56.00  $\pm$ 4.25 {\color{olivegreen}(57.18)} &           31.04 $\pm$ 16.65{\color{olivegreen}(33.77)} &           92.94  $\pm$ 0.66  {\color{olivegreen}(93.38)} &           84.63  $\pm$ 0.17 {\color{olivegreen}(85.47)}  \\
                               &  HiMatch                                        ${\dagger}$   &           57.43  $\pm$ 0.01 {\color{olivegreen}(57.43)} &           39.04 $\pm$ 0.01 {\color{olivegreen}(39.04)} &                                -                         &                                -                         \\
                               &  HGCLR                                          ${\dagger}$   &           56.80  $\pm$ 4.24 {\color{olivegreen}(57.96)} &           32.34 $\pm$ 15.39{\color{olivegreen}(33.76)} &           93.14  $\pm$ 0.01  {\color{olivegreen}(93.22)} &           84.74  $\pm$ 0.11 {\color{olivegreen}(85.11)}  \\
                               &  HPT                                            ${\dagger}$   &           65.57  $\pm$ 1.69 {\color{olivegreen}(67.06)} &           45.89 $\pm$ 9.78 {\color{olivegreen}(49.42)} &           94.34  $\pm$ 0.28  {\color{olivegreen}(94.83)} &           90.09  $\pm$ 0.87 {\color{olivegreen}(91.12)}  \\
                               &  HierVerb                                       ${\dagger}$   &           72.58  $\pm$ 0.83 {\color{olivegreen}(73.64)} &           63.12 $\pm$ 1.48 {\color{olivegreen}(64.47)} &           94.75  $\pm$ 0.13  {\color{olivegreen}(95.13)} &           90.77  $\pm$ 0.33 {\color{olivegreen}(91.43)}  \\
                               &  EPR                                                          &           38.42  $\pm$ 0.91 {\color{olivegreen}(39.36)} &           19.94 $\pm$ 1.32 {\color{olivegreen}(21.31)} &           27.94  $\pm$ 1.47  {\color{olivegreen}(29.56)} &           18.31  $\pm$ 1.70 {\color{olivegreen}(20.09)}  \\
                               &  REGEN                                                        &           58.75  $\pm$ 2.04 {\color{olivegreen}(60.71)} &           33.20 $\pm$ 2.01 {\color{olivegreen}(35.40)} &           94.11  $\pm$ 0.79  {\color{olivegreen}(95.01)} &           86.76  $\pm$ 1.04 {\color{olivegreen}(87.92)}  \\  
                               \cmidrule{2-6}
                               &                    Retrieval                                  &\underline{75.37} $\pm$ 0.70 {\color{olivegreen}(76.08)} &\underline{65.94}$\pm$ 0.57 {\color{olivegreen}(66.41)} &\underline{95.15} $\pm$ 0.07  {\color{olivegreen}(95.23)} &\underline{91.26} $\pm$ 0.14 {\color{olivegreen}(91.38)}  \\  
                               &              Retrieval-style ICL                              &\textbf{   75.62  $\pm$ 0.15 {\color{olivegreen}(75.78)}}&\textbf{   66.34 $\pm$ 0.09 {\color{olivegreen}(66.41)}}&\textbf{   95.26  $\pm$ 0.07  {\color{olivegreen}(95.23)}}&\textbf{   91.42  $\pm$ 0.05 {\color{olivegreen}(91.47)}} \\ \midrule

            \multirow{9}{*}{8} &              BERT                               ${\dagger}$   &           66.24  $\pm$ 1.96 {\color{olivegreen}(67.53)} &           50.21 $\pm$ 5.05 {\color{olivegreen}(52.60)} &           94.39  $\pm$ 0.06  {\color{olivegreen}(94.57)} &           87.63  $\pm$ 0.28 {\color{olivegreen}(87.78)}  \\
                               &  HiMatch                                        ${\dagger}$   &           69.92  $\pm$ 0.01 {\color{olivegreen}(70.23)} &           57.47 $\pm$ 0.01 {\color{olivegreen}(57.78)} &                                 -                        &                                -                         \\
                               &  HGCLR                                          ${\dagger}$   &           68.34  $\pm$ 0.96 {\color{olivegreen}(69.22)} &           54.41 $\pm$ 2.97 {\color{olivegreen}(55.99)} &           94.70  $\pm$ 0.05  {\color{olivegreen}(94.94)} &           88.04  $\pm$ 0.25 {\color{olivegreen}(88.61)}  \\
                               &  HPT                                            ${\dagger}$   &           76.22  $\pm$ 0.99 {\color{olivegreen}(77.23)} &           67.20 $\pm$ 1.89 {\color{olivegreen}(68.63)} &           95.49  $\pm$ 0.01  {\color{olivegreen}(95.57)} &           92.35  $\pm$ 0.03 {\color{olivegreen}(92.52)}  \\
                               &  HierVerb                                       ${\dagger}$   &\underline{78.12} $\pm$ 0.55 {\color{olivegreen}(78.87)} &\underline{69.98}$\pm$ 0.91 {\color{olivegreen}(71.04)} &\underline{95.69} $\pm$ 0.01  {\color{olivegreen}(95.70)} &\underline{92.44} $\pm$ 0.01 {\color{olivegreen}(92.51)}  \\
                               &  EPR                                                          &           41.35  $\pm$ 0.43 {\color{olivegreen}(41.83)} &           22.19 $\pm$ 0.32 {\color{olivegreen}(22.57)} &           44.95  $\pm$ 0.43  {\color{olivegreen}(45.42)} &           31.13  $\pm$ 0.38 {\color{olivegreen}(31.56)}  \\
                               &  REGEN                                                        &           67.91  $\pm$ 1.47 {\color{olivegreen}(69.54)} &           55.39 $\pm$ 1.86 {\color{olivegreen}(57.32)} &           95.24  $\pm$ 0.12  {\color{olivegreen}(95.38)} &           90.56  $\pm$ 0.39 {\color{olivegreen}(90.99)} \\  
                               \cmidrule{2-6}
                               &                    Retrieval                                  &\textbf{   79.04  $\pm$ 0.48 {\color{olivegreen}(79.53)}}&\textbf{   70.59 $\pm$ 0.52 {\color{olivegreen}(71.04)}}&\textbf{   95.71  $\pm$ 0.06  {\color{olivegreen}(95.78)}}&\textbf{   92.50  $\pm$ 0.02 {\color{olivegreen}(92.52)}} \\  
                               &              Retrieval-style ICL                              &           76.93  $\pm$ 0.05 {\color{olivegreen}(76.98)} &           67.54 $\pm$ 0.04 {\color{olivegreen}(67.57)} &           95.43  $\pm$ 0.01  {\color{olivegreen}(95.44)} &           91.85  $\pm$ 0.01 {\color{olivegreen}(91.86)}  \\ \midrule

            \multirow{9}{*}{16}&              BERT                               ${\dagger}$   &           75.52  $\pm$ 0.32 {\color{olivegreen}(76.07)} &           65.85 $\pm$ 1.28 {\color{olivegreen}(66.96)} &           95.31  $\pm$ 0.01  {\color{olivegreen}(95.37)} &           89.16  $\pm$ 0.07 {\color{olivegreen}(89.35)}  \\
                               &  HiMatch                                        ${\dagger}$   &           77.67  $\pm$ 0.01 {\color{olivegreen}(78.24)} &           68.70 $\pm$ 0.01 {\color{olivegreen}(69.58)} &                                 -                        &                                -                         \\
                               &  HGCLR                                          ${\dagger}$   &           76.93  $\pm$ 0.52 {\color{olivegreen}(77.46)} &           67.92 $\pm$ 1.21 {\color{olivegreen}(68.66)} &           95.49  $\pm$ 0.04  {\color{olivegreen}(95.63)} &           89.41  $\pm$ 0.09 {\color{olivegreen}(89.71)}  \\
                               &  HPT                                            ${\dagger}$   &           79.85  $\pm$ 0.41 {\color{olivegreen}(80.58)} &           72.02 $\pm$ 1.40 {\color{olivegreen}(73.31)} &           96.13  $\pm$ 0.01  {\color{olivegreen}(96.21)} &\underline{93.34} $\pm$ 0.02 {\color{olivegreen}(93.45)}  \\
                               &  HierVerb                                       ${\dagger}$   &\underline{80.93} $\pm$ 0.10 {\color{olivegreen}(81.26)} &\textbf{   73.80 $\pm$ 0.12 {\color{olivegreen}(74.19)}}&\underline{96.17} $\pm$ 0.01  {\color{olivegreen}(96.21)} &           93.28  $\pm$ 0.06 {\color{olivegreen}(93.49)}  \\  
                               &  EPR                                                          &           44.57  $\pm$ 0.09 {\color{olivegreen}(44.70)} &           24.50 $\pm$ 0.18 {\color{olivegreen}(24.74)} &           52.68  $\pm$ 0.04  {\color{olivegreen}(52.71)} &           42.76  $\pm$ 0.03 {\color{olivegreen}(42.78)}  \\
                               &  REGEN                                                        &           77.64  $\pm$ 1.04 {\color{olivegreen}(78.70)} &           69.91 $\pm$ 1.68 {\color{olivegreen}(71.68)} &           95.88  $\pm$ 0.03  {\color{olivegreen}(95.91)} &           91.73  $\pm$ 0.07 {\color{olivegreen}(91.80)} \\  
                               \cmidrule{2-6}
                               &                    Retrieval                                  &\textbf{   81.12  $\pm$ 0.26 {\color{olivegreen}(81.38)}}&\underline{73.72}$\pm$ 0.17 {\color{olivegreen}(73.82)} &\textbf{   96.22  $\pm$ 0.04  {\color{olivegreen}(96.27)}}&\textbf{   93.37  $\pm$ 0.02 {\color{olivegreen}(93.46)}} \\
                               &              Retrieval-style ICL                              &           78.62  $\pm$ 0.03 {\color{olivegreen}(78.65)} &           69.56 $\pm$ 0.03 {\color{olivegreen}(69.59)} &           95.56  $\pm$ 0.00  {\color{olivegreen}(95.56)} &           92.04  $\pm$ 0.00 {\color{olivegreen}(92.04)}  \\

        \bottomrule
    \end{tabular}
    \caption{Micro-F1 and Macro-F1 scores on two English datasets. We reported the average, standard deviation, and best results across three random seeds. \textbf{Bold}: the best result. \underline{Underlined}: the second highest. ${\dagger}$: the direct utilization of results from \citet{ji-etal-2023-hierarchical}.}
    \label{tab:main_result}
    \end{table*}

\begin{table}[t]
    \centering
    \resizebox{1\columnwidth}{!}{
        \begin{tabular}{c l c c}
            \toprule
            
            \multirow{2}{*}{Q} & \makecell[c]{\multirow{2}{*}{Method}} & \multicolumn{2}{c}{\textbf{Patent(Depth 4)}} \\
            \cmidrule{3-4}
            & & Micro-F1 & Macro-F1 \\
            \midrule
            \multirow{6}{*}{1} &  BERT                &            27.04  $\pm$ 1.48  {\color{olivegreen}(28.37)} &           2.40   $\pm$ 0.23 {\color{olivegreen}(2.67)}   \\
                               &  HGCLR               &            28.99  $\pm$ 1.12  {\color{olivegreen}(29.89)} &           2.94   $\pm$ 0.24 {\color{olivegreen}(3.13)}   \\
                               &  HPT                 &            35.22  $\pm$ 1.07  {\color{olivegreen}(36.26)} &           4.22   $\pm$ 0.68 {\color{olivegreen}(4.87)}   \\
                               &  HierVerb            &            41.83  $\pm$ 0.60  {\color{olivegreen}(42.52)} &           5.91   $\pm$ 0.65 {\color{olivegreen}(6.53)}   \\  \cmidrule{2-4}
                               &  Retrieval           & \underline{47.71} $\pm$ 0.41  {\color{olivegreen}(48.15)} &\underline{10.76} $\pm$ 0.25 {\color{olivegreen}(11.02)}  \\
                               &  Retrieval-style ICL & \textbf{   52.23  $\pm$ 0.23  {\color{olivegreen}(52.45)}}&\textbf{   15.62  $\pm$ 0.17 {\color{olivegreen}(15.78)}} \\  \midrule

            \multirow{6}{*}{2} &  BERT                &            36.07  $\pm$ 0.24  {\color{olivegreen}(36.88)} &           6.41   $\pm$ 0.77 {\color{olivegreen}(6.92)}   \\
                               &  HGCLR               &            36.73  $\pm$ 1.13  {\color{olivegreen}(38.03)} &           6.82   $\pm$ 0.21 {\color{olivegreen}(7.06)}   \\
                               &  HPT                 &            42.61  $\pm$ 0.75  {\color{olivegreen}(43.43)} &           10.53  $\pm$ 0.30 {\color{olivegreen}(10.87)}  \\
                               &  HierVerb            &            48.42  $\pm$ 0.39  {\color{olivegreen}(48.74)} &           12.97  $\pm$ 0.39 {\color{olivegreen}(13.24)}  \\  \cmidrule{2-4}
                               &  Retrieval           & \underline{51.63} $\pm$ 0.34  {\color{olivegreen}(51.91)} &\underline{15.12} $\pm$ 0.23 {\color{olivegreen}(15.32)}  \\  
                               &  Retrieval-style ICL & \textbf{   56.84  $\pm$ 0.11  {\color{olivegreen}(56.93)}}&\textbf{   20.07  $\pm$ 0.10 {\color{olivegreen}(20.17)}} \\  \midrule
            
            \multirow{6}{*}{4} &  BERT                &            49.41  $\pm$ 0.98  {\color{olivegreen}(50.24)} &           9.64   $\pm$ 0.56 {\color{olivegreen}(10.13)}  \\
                               &  HGCLR               &            50.24  $\pm$ 0.36  {\color{olivegreen}(50.63)} &           11.40  $\pm$ 0.29 {\color{olivegreen}(11.67)}  \\
                               &  HPT                 &            53.91  $\pm$ 0.44  {\color{olivegreen}(54.29)} &           18.45  $\pm$ 0.40 {\color{olivegreen}(18.79)}  \\
                               &  HierVerb            &            57.58  $\pm$ 0.83  {\color{olivegreen}(58.64)} &           23.28  $\pm$ 0.39 {\color{olivegreen}(23.63)}  \\  \cmidrule{2-4}
                               &  Retrieval           & \textbf{   60.53  $\pm$ 0.36  {\color{olivegreen}(60.82)}}&\textbf{   25.65  $\pm$ 0.29 {\color{olivegreen}(25.95)}} \\  
                               &  Retrieval-style ICL & \underline{59.35} $\pm$ 0.10  {\color{olivegreen}(59.45)} &\underline{24.15} $\pm$ 0.08 {\color{olivegreen}(24.25)}  \\  \midrule

            \multirow{6}{*}{8} &  BERT                &            62.10  $\pm$ 1.34  {\color{olivegreen}(63.29)} &           26.85  $\pm$ 0.97 {\color{olivegreen}(27.75)}  \\
                               &  HGCLR               &            64.69  $\pm$ 0.30  {\color{olivegreen}(65.01)} &           27.69  $\pm$ 0.47 {\color{olivegreen}(28.21)}  \\
                               &  HPT                 &            67.35  $\pm$ 0.13  {\color{olivegreen}(67.45)} &           28.39  $\pm$ 0.08 {\color{olivegreen}(28.46)}  \\
                               &  HierVerb            & \underline{68.74} $\pm$ 0.12  {\color{olivegreen}(68.82)} &\underline{29.93} $\pm$ 0.07 {\color{olivegreen}(30.01)}  \\  \cmidrule{2-4}
                               &  Retrieval           & \textbf{   69.44  $\pm$ 0.10  {\color{olivegreen}(69.53)}}&\textbf{   30.32  $\pm$ 0.07 {\color{olivegreen}(30.38)}} \\  
                               &  Retrieval-style ICL &            65.81  $\pm$ 0.05  {\color{olivegreen}(65.86)} &           27.86  $\pm$ 0.04 {\color{olivegreen}(27.89)}  \\  \midrule

            \multirow{6}{*}{16}&  BERT                &            70.97  $\pm$ 0.36  {\color{olivegreen}(71.32)} &           30.90  $\pm$ 0.39 {\color{olivegreen}(31.34)}  \\
                               &  HGCLR               &            71.44  $\pm$ 0.38  {\color{olivegreen}(71.74)} &           31.87  $\pm$ 0.05 {\color{olivegreen}(31.85)}  \\
                               &  HPT                 &            73.23  $\pm$ 0.17  {\color{olivegreen}(73.37)} &           33.44  $\pm$ 0.17 {\color{olivegreen}(33.60)}  \\
                               &  HierVerb            & \underline{75.72} $\pm$ 0.11  {\color{olivegreen}(75.85)} &\underline{34.75} $\pm$ 0.10 {\color{olivegreen}(34.86)}  \\  \cmidrule{2-4}
                               &  Retrieval           & \textbf{   75.94  $\pm$ 0.11  {\color{olivegreen}(76.06)}}&\textbf{   34.95  $\pm$ 0.05 {\color{olivegreen}(35.00)}} \\
                               &  Retrieval-style ICL &            68.73  $\pm$ 0.03  {\color{olivegreen}(68.76)} &           29.82  $\pm$ 0.03 {\color{olivegreen}(29.85)}  \\
        \bottomrule
    \end{tabular}
    }
    \caption{Micro-F1 and Macro-F1 scores on the Chinese Patent dataset. We reported the average, standard deviation, and best results across three random seeds. \textbf{Bold}: the best result. \underline{Underlined}: the second highest.}
    \label{tab:main_result_cn}
    \end{table}

\textbf{Retrieval} employs our retrieval method to select the label associated with the text in the retrieval database that has the highest similarity score as the label for the test text.

\textbf{Retrieval-style ICL} involves the selection of the top three (K$=$3) documents with distinct labels from the retrieval database. Subsequently, these documents and labels are utilized as demonstrations to construct the prompt, and our iterative method is applied to hierarchical label inference.

It is worth mentioning that in our LLM generative approach, if the generated label is not present in the candidate label set, the label corresponding to the retrieval text with the highest similarity is selected as its inference result.

\subsection{Main Results}
The main results are shown in Table \ref{tab:main_result} and Table \ref{tab:main_result_cn}.
It can be observed that our retrieval-based approach achieved the best results across almost all settings.
Also, we find that our method is less affected by random seeds, resulting in more stable and robust performance, which further demonstrated the effectiveness of our approach.

Specifically, as Q increases, all methods improve continuously.
However, our retrieval-based method consistently performs the best, and its advantages become even more pronounced in extremely low-resource settings.
In the 1-shot setting, compared with the previous state-of-the-art model, \texttt{Retrieval} shows an average of $4.51\%$ micro, $5.28\%$ macro-F1 absolute improvement on WOS, $1.87\%$ micro, $3.09\%$ macro-F1 absolute on DBPedia, and $5.88\%$ micro-F1, $4.85\%$ macro-F1 absolute improvement on Patent.
As the hierarchy depth increases, we find that /texttt[Retrieval] exhibits an advantage even in the 16-shot setting.
We think it is because label text descriptions better differentiate categories, especially in the deeper HTC dataset.
Despite the fact that our method is still the most effective, we observe that all methods in Patent don't perform well due to the deep hierarchy and the large number of labels.
Furthermore, by examining the results on the Patent dataset, we observe that all methods exhibit similar trends to those on the English dataset, which also confirms the effectiveness of hierarchical classification approaches for Chinese HTC tasks.
In the 1-shot, 2-shot, and 4-shot settings, \texttt{Retrieval-style ICL} achieves outstanding performance.

The EPR also uses a retrieval strategy.
Following \citet{rubin-etal-2022-learning}, we replicate its model on the HTC task\footnote{The scoring LM utilized by EPR is GPT-neo, which does not perform well on Chinese. Therefore, we only present experimental results conducted on English datasets.},
and it demonstrated inferior performance compared to our method.
There are probably two factors causing this performance gap. On the one hand, EPR uses the poorly performing GPT-neo 2.7B as the scoring and inference LM and does not fine-tune it during the training process.
Especially in the few shot setting, the ability of scoring LM itself has a significant impact on the experimental results.
On the other hand, EPR is not proposed for HTC.
Therefore, it does not utilize the information of hierarchical relationship between labels, which leads to a mismatch between the retrieved samples and the target sample.
Furthermore, based on our observations, we find that the performance of EPR on the simpler dataset DBpedia is even inferior to that on WOS when compared to other methods.
This discrepancy could be attributed to the deeper hierarchy of DBpedia, which leads to a larger number of labels and increases retrieval difficulty.
In contrast, our proposed method incorporates classification objective loss and leverages hierarchical label information, which remains unaffected by these challenges and ensures more robust performance.

Observing Table \ref{tab:main_result}, it can be observed that REGEN results, particularly in terms of Macro-F1, exhibit performance gaps compared to other HTC methods.
This discrepancy stems from the fact that REGEN does not utilize label hierarchy information and focuses only on predicting leaf nodes.
It reflects the significance of considering the label hierarchy structure of HTC.


\subsection{Analysis}
\label{section:Analysis}

\begin{figure}[t]
    \centering
    \includegraphics[width=1\columnwidth]{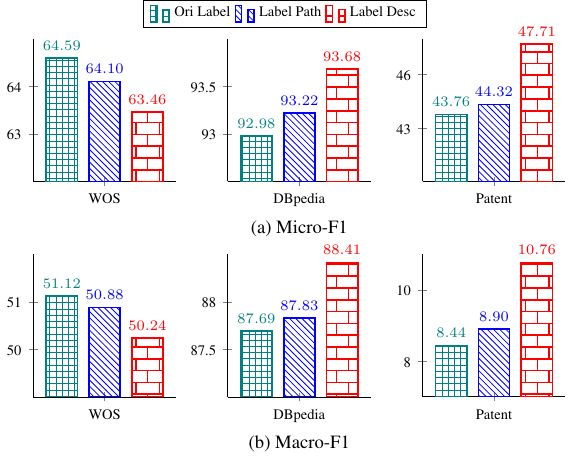}
    \caption{Results of different label text types in the 1-shot setting. \texttt{Ori Label} means the original leaf label text, \texttt{Label Path} means all text on the label path, and \texttt{Label Desc} means the label description text of LLM.}
    \label{fig:label_description}
\end{figure}

\paragraph{The impact of label descriptions on retrieval.}
We conduct experiments on different types of label texts:
(1) original leaf label text,
(2) all text on the label path,
and (3) label descriptions generated by LLM.
The results are shown in Figure \ref{fig:label_description}.

We find that on WOS, (1)$>$(2)$>$(3), while on DBpedia and Patent datasets, (1)$<$(2)$<$(3).
We analyze that it may be due to the shallow hierarchy and small number of labels in the WOS dataset.
The label text itself has a high degree of discrimination, so adding additional information leads to a decrease.
In contrast, the deeper hierarchy and larger number of labels in the DBpedia and Patent datasets require more information to distinguish the semantic meaning of label text.
The experiment proves that label text improves retrieval results. However, which type of label text to use needs to be selected according to the dataset.

\paragraph{Comparison with different contrastive learning strategies.}
To demonstrate the effectiveness of our divergent contrastive learning, we illustrate the results with three more straightforward losses.
\texttt{CL Hierarchical} denotes we calculate $\mathcal{L}_{con}$ for each hierarchical label representation with random sampling among a batch.
\texttt{CL Leaf Only} refers we only calculate loss between leaf label representations with random sampling among a batch.
\texttt{w/o CL} means training without $\mathcal{L}_{con}$ in Equation \ref{eq:loss}.

\usepgfplotslibrary{groupplots}
\usetikzlibrary{patterns,backgrounds}

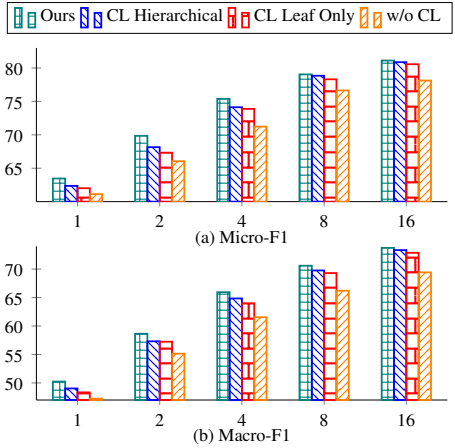
\begin{figure}[t]
\centering
\resizebox{0.8\columnwidth}{!}{
\begin{tikzpicture}
    \pgfkeys{/pgf/number format/.cd,fixed, fixed zerofill,precision=2}
	\begin{groupplot}[group style={group name=myplot,group size=1 by 2,horizontal sep=25pt,vertical sep=20pt,xlabels at=edge bottom, ylabels at=edge left},height=8cm,width=8cm]

    \nextgroupplot[
    ybar,
    bar width=10pt,
    xtick={1,2,3,4,5},
    xticklabels={1,2,4,8,16},
    xmax=5.5,xmin=0.5,
    x tick label style = {yshift=-0.5em, text height=0ex,font=\scriptsize},
    ytick = {65, 70, 75, 80},
    yticklabels={65, 70, 75, 80},
    ymax=83,ymin=60,
    y tick label style = {yshift=-0.3em, text height=0ex,font=\scriptsize},
	x label style = {font=\scriptsize},
	y label style = {font=\scriptsize},
    axis x line = bottom,
    axis y line=left,
    width = 8cm,
    height = 4cm,
    axis line style={-},
    title style={xshift=0em,yshift=-9em,font=\scriptsize},
    title=(a) Micro-F1,
    legend style={xshift=-8.5em, yshift=2em, text height=0ex, anchor=north, legend columns=4, font=\scriptsize},
    ]

    \addplot[teal, pattern=grid, thick, pattern color=teal, bar shift=-0.75em, bar width = 0.5em] coordinates
{
    (1,  63.46)
    (2,  69.85)
    (3,  75.37)
    (4,  79.04)
    (5,  81.12)
};\addlegendentry{Ours}
\addplot[blue, pattern= north west lines, pattern color=blue, thick, bar shift=-0.25em, bar width = 0.5em] coordinates
{
    (1,  62.36)
    (2,  68.15)
    (3,  74.13)
    (4,  78.84)
    (5,  80.87)
};\addlegendentry{CL Hierarchical}
\addplot[red, pattern=bricks, thick, pattern color=red, bar shift=0.25em, bar width = 0.5em] coordinates
{
    (1,  62.01)
    (2,  67.33)
    (3,  73.88)
    (4,  78.31)
    (5,  80.56)
};\addlegendentry{CL Leaf Only}
\addplot[orange, pattern=north east lines, thick, pattern color=orange, bar shift=0.75em, bar width = 0.5em] coordinates
{
    (1,  61.12)
    (2,  66.05)
    (3,  71.23)
    (4,  76.64)
    (5,  78.12)
};\addlegendentry{w/o CL}

\nextgroupplot[
    ybar,
    bar width=10pt,
    xtick={1,2,3,4,5},
    xticklabels={1,2,4,8,16},
    xmax=5.5,xmin=0.5,
    x tick label style = {yshift=-0.5em, text height=0ex,font=\scriptsize},
    ytick = {50, 55, 60, 65, 70},
    yticklabels={50, 55, 60, 65, 70},
    ymax=74,ymin=47,
    y tick label style = {yshift=-0.3em, text height=0ex,font=\scriptsize},
	x label style = {font=\scriptsize},
	y label style = {font=\scriptsize},
    axis x line = bottom,
    axis y line=left,
    width = 8cm,
    height = 4cm,
    axis line style={-},
    title style={xshift=0em,yshift=-9em,font=\scriptsize},
    title=(b) Macro-F1,
    legend style={xshift=-12em, yshift=3em, text height=0ex, anchor=north, legend columns=4, font=\scriptsize},
    ]

    \addplot[teal, pattern=grid, thick, pattern color=teal, bar shift=-0.75em, bar width = 0.5em] coordinates
{
    (1,  50.24)
    (2,  58.64)
    (3,  65.94)
    (4,  70.59)
    (5,  73.72)
};
\addplot[blue, pattern= north west lines, pattern color=blue, thick, bar shift=-0.25em, bar width = 0.5em] coordinates
{
    (1,  49.03)
    (2,  57.31)
    (3,  64.84)
    (4,  69.76)
    (5,  73.33)
};
\addplot[red, pattern=bricks, thick, pattern color=red, bar shift=0.25em, bar width = 0.5em] coordinates
{
    (1,  48.34)
    (2,  57.24)
    (3,  64.00)
    (4,  69.30)
    (5,  72.86)
};
\addplot[orange, pattern=north east lines, thick, pattern color=orange, bar shift=0.75em, bar width = 0.5em] coordinates
{
    (1,  47.24)
    (2,  55.14)
    (3,  61.54)
    (4,  66.20)
    (5,  69.42)
};

\end{groupplot}

\end{tikzpicture}
}
\caption{Results of different contrastive learning strategy on WOS dataset. The x-axis denotes the shot number Q and the y-axis denotes the F1 score.}\label{fig:loss}
\end{figure}

As shown in Figure \ref{fig:loss}, our divergent contrastive learning outperforms the others through all the shot numbers.
Previous research has shown that contrastive learning is an effective option for training dense retrievers \cite{DBLP:conf/sigir/0001SL23, DBLP:conf/iclr/XiongXLTLBAO21}.
\texttt{w/o CL} has the lowest performance compared to other contrastive learning methods.
As opposed to \texttt{CL Leaf Only} that treats HTC as a flat classification, \texttt{CL Hierarchical} models the label path information.
Our divergent contrast learning selects more hard negative samples based on label similarity, further spatially pulling apart the vector distribution of samples with similar labels.

\paragraph{Comparison between classification-based and retrieval-based methods.}
Previous research on HTC has mainly used classification-based methods,
which train classifiers to predict the probability distribution of each label.
In contrast, our proposed retrieval-based method predicts labels by calculating similarity with a retrieval database to obtain the most similar text and corresponding labels.
Therefore, we replaced our retrieval prediction with classifier prediction while keeping other settings consistent, and compared classification-based and retrieval-based methods.
The results are shown in Table \ref{tab:classification_retrieve}.

\begin{table}[t]
    \centering
    \resizebox{1\columnwidth}{!}{
        \begin{tabular}{cc | cc }
            \toprule



            \textbf{Q} & & \textbf{Classification} & \textbf{Retrieval} \\ \midrule

            \multirow{2}{*}{1}  & Micro-F1 &         63.25 $\pm$ 2.17 {\color{olivegreen}(65.61)}  & \textbf{63.46 $\pm$ 2.30 {\color{olivegreen}(65.99)}}  \\
                                & Macro-F1 &         49.91 $\pm$ 2.43 {\color{olivegreen}(52.65)}  & \textbf{50.24 $\pm$ 2.21 {\color{olivegreen}(52.66)}}  \\ \midrule
            \multirow{2}{*}{2}  & Micro-F1 &         69.09 $\pm$ 0.57 {\color{olivegreen}(69.74)}  & \textbf{69.85 $\pm$ 0.63 {\color{olivegreen}(70.58)}}  \\
                                & Macro-F1 &         58.49 $\pm$ 0.46 {\color{olivegreen}(59.04)}  & \textbf{58.64 $\pm$ 0.58 {\color{olivegreen}(59.25)}}  \\ \midrule
            \multirow{2}{*}{4}  & Micro-F1 &         74.48 $\pm$ 0.74 {\color{olivegreen}(75.34)}  & \textbf{75.37 $\pm$ 0.70 {\color{olivegreen}(76.08)}}  \\
                                & Macro-F1 &         65.78 $\pm$ 0.60 {\color{olivegreen}(66.36)}  & \textbf{65.94 $\pm$ 0.57 {\color{olivegreen}(66.41)}}  \\ \midrule
            \multirow{2}{*}{8}  & Micro-F1 &         78.36 $\pm$ 0.15 {\color{olivegreen}(78.48)}  & \textbf{79.04 $\pm$ 0.48 {\color{olivegreen}(79.53)}}  \\
                                & Macro-F1 &         70.55 $\pm$ 0.34 {\color{olivegreen}(70.93)}  & \textbf{70.59 $\pm$ 0.52 {\color{olivegreen}(71.04)}}  \\ \midrule
            \multirow{2}{*}{16} & Micro-F1 &         80.92 $\pm$ 0.21 {\color{olivegreen}(81.06)}  & \textbf{81.12 $\pm$ 0.26 {\color{olivegreen}(81.38)}}  \\
                                & Macro-F1 & \textbf{73.88 $\pm$ 0.21 {\color{olivegreen}(74.08)}} &         73.72 $\pm$ 0.17 {\color{olivegreen}(73.82)}   \\ 
        \bottomrule
    \end{tabular}
    }
    \caption{The results of classification-based and retrieval-based methods on WOS dataset. We reported the average, standard deviation, and best results across three random seeds. \textbf{Bold}: the best result.}
    \label{tab:classification_retrieve}
    \end{table}

\begin{table}[t]
    \centering
    \resizebox{1\columnwidth}{!}{
        \begin{tabular}{lccccc}
            \toprule

            \multirow{3}{*}{\textbf{Method}} & \multirow{3}{*}{\textbf{Level}} & \multicolumn{4}{c}{\textbf{DBpedia}} \\
            \cmidrule{3-6}
                                & & \multicolumn{2}{c}{\textbf{Q=1}} & \multicolumn{2}{c}{\textbf{Q=16}} \\
                                & & Micro-F1 & Macro-F1 & Micro-F1 & Macro-F1 \\
                                \midrule

            \multirow{3}{*}{BERT} & 1 &  17.60 &  5.13 & 98.42 & 94.55 \\
                                  & 2 &      14.02     &       0.31      &      93.81      &       90.69    \\
                                  & 3 &      11.17     &       0.10      &      90.13      &       85.92    \\
                                \midrule
                                \multirow{3}{*}{EPR} 
                                  & 1 &      23.51     &       11.51     &      71.40      &       69.51    \\
                                  & 2 &      8.59      &       8.38      &      53.71      &       48.74    \\
                                  & 3 &      7.06      &       6.51      &      43.67      &       39.71    \\
                                \midrule
                                \multirow{3}{*}{Retrieval}
                                  & 1 &      98.34     &       95.08     &       98.75     &       96.58    \\
                                  & 2 &      93.05     &       89.85     &       94.93     &       91.17    \\
                                  & 3 &      89.14     &       87.62     &       90.66     &       87.01    \\
        \bottomrule
    \end{tabular}
    }
    \caption{Results at different hierarchy levels on DBpedia dataset.}
    \label{tab:layer_compare}
    \end{table}

We find that under the few-shot setting, the retrieval-based method outperformed the classification-based method, although the gap gradually decreased with an increasing number of training samples.
We speculate that in settings with a small number of samples, the classifier may not be well-trained, while index vectors generated during the retrieval process have better semantic representations due to the rich semantic knowledge of pre-trained models.
The retrieval-based method that utilizes similarity matching can achieve relatively better performance, especially in terms of Micro-F1.

Concurrently, we compare the classic classification method BERT, the retrieval-based method EPR, and our method on DBpedia which has a deeper hierarchy.
Table \ref{tab:layer_compare} illustrates the performance of these three methods at different hierarchical levels.
It is observed that BERT performs well with slightly more samples, EPR excels with extremely limited samples, and our method consistently demonstrates excellent performance across scenarios.

\paragraph{Impact of imbalanced few-shot sampling}
We sample few-shot training set with Algorithm \ref{algorithm:sample}, where we enforce balanced control over each type of samples.
We now explore the impact of a bias training set and replace the step 13 to 17 in Algorithm \ref{algorithm:sample} to:
\begin{align*}
    \text{SampleN} = \text{Random}(0, \text{Min}(\mathcal{C}[y_i], Q))\;\\
    \mathcal{S} = \mathcal{S} \cup \text{Random Sample}(\mathcal{C}[y_i], \text{SampleN}).\;
\end{align*}

\begin{table}[t]
    \centering
    \resizebox{1\columnwidth}{!}{
        \begin{tabular}{cl cc }
            \toprule
            \multirow{2}{*}{\textbf{Q}}          &                          & \multicolumn{2}{c}{\textbf{WOS}}  \\ \cmidrule{3-4}
                                &               &             Micro-F1     &         Macro-F1                  \\ \midrule

            \multirow{7}{*}{16} &  BERT               &           70.42  $\pm$ 3.43 {\color{olivegreen}(74.07)}  &           57.38 $\pm$ 7.98 {\color{olivegreen}(65.36)}  \\
                                &  HiMatch            &           72.67  $\pm$ 4.97 {\color{olivegreen}(77.64)}  &           61.80 $\pm$ 7.89 {\color{olivegreen}(69.80)}  \\
                                &  HGCLR              &           71.93  $\pm$ 4.48 {\color{olivegreen}(76.41)}  &           60.72 $\pm$ 5.83 {\color{olivegreen}(67.63)}  \\
                                &  HPT                &           73.85  $\pm$ 4.33 {\color{olivegreen}(78.18)}  &           65.02 $\pm$ 6.70 {\color{olivegreen}(73.39)} \\
                                &  HierVerb           &           75.63  $\pm$ 3.80 {\color{olivegreen}(79.62)}  &           64.77 $\pm$ 7.60 {\color{olivegreen}(73.39)}  \\  
                                &  EPR                &           39.66  $\pm$ 5.17 {\color{olivegreen} (44.90)} &           15.67 $\pm$ 7.10 {\color{olivegreen} (23.13)} \\
                                &  Retrieval          &\textbf{   79.42  $\pm$ 3.82 {\color{olivegreen}(83.81)}} & \textbf{  69.02 $\pm$ 5.56 {\color{olivegreen}(74.92)}} \\

        \bottomrule
    \end{tabular}
    }
    \caption{Results of randomly sampled 16-shot setting on WOS.}
    \label{tab:sampling}
    \end{table}

The report the results on WOS dataset in Table \ref{tab:sampling}.
We observe that the average F1 values of all the methods decrease and the random sampling approach makes a wider range of results.
Our method could keep in lead and delivers a more stable performance.

\begin{figure}[t]
    \centering
    \includegraphics[width=1\columnwidth]{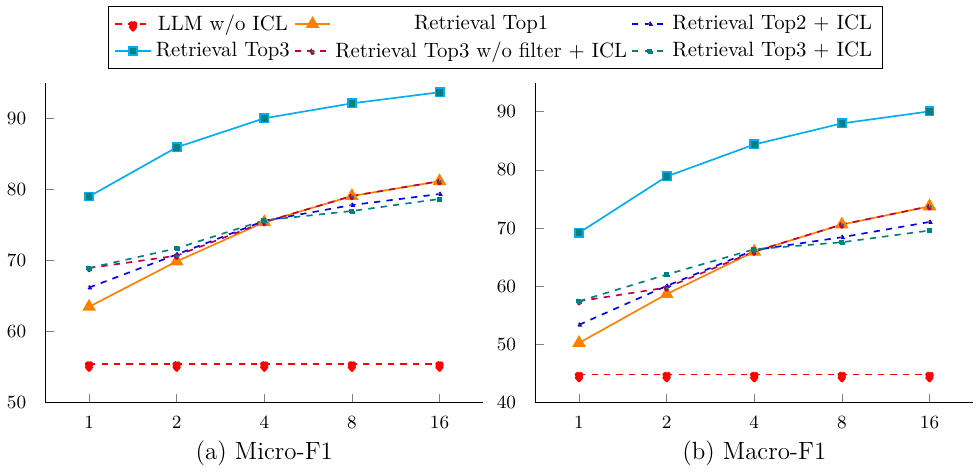}
    \caption{Micro-F1 (a) and Macro-F1 (b) results curves of top k retrieval and ICL with different numbers of examples on WOS. The horizontal axis represents the number of shots in the training set, and the vertical axis represents the metric value (\%). \texttt{w/o} means 'without'.}
    \label{fig:ICL}
\end{figure}

\paragraph{Comparison with zero-shot setting on LLM.}
The quality of instances in the ICL prompt directly affects the results of ICL inference.
We compared the impact of retrieval on ICL inference under different settings, and the results are shown in Figure \ref{fig:ICL}.
We distinguish between different methods using different lines, where the soild line represents the retrieval-based methods and the dashed line represents the LLM-based methods.

\texttt{LLM w/o ICL} refers to the situation where no examples are provided,
and only the test document and the label set are given to the large model for inference.
In other words, under the zero-shot setting, the inference relies entirely on the strong ability of LLM.
Due to the large and complex label space, it is difficult to input it to the large model for inference at once.
Therefore, \texttt{LLM w/o ICL} also uses iterative inference, sequentially inputting the label corresponding sub-clusters.
We find that even under zero-shot setting, the large model still demonstrate strong performance,
with $55.40\%$ Micro-F1 and $44.79\%$ Macro-F1,
which even outperform classification results of vanilla fine-tuned BERT under 4-shot.

\paragraph{Comparison with different LLM base models.}
\begin{table}[t]
    \centering
    \resizebox{1\columnwidth}{!}{
        \begin{tabular}{cl | cc }
            \toprule



            \textbf{Q} & \makecell[c]{\textbf{Model}} & \textbf{Micro-F1} & \textbf{Macro-F1} \\ \midrule

            \multirow{4}{*}{1}  & BERT (Vanilla FT)     &            2.99  $\pm$ 20.85 {\color{olivegreen}(5.12)}   &            0.16  $\pm$ 0.10 {\color{olivegreen}(0.24)}    \\
                                & Llama-7B (Seq2Seq FT) &            42.76 $\pm$ 1.30 {\color{olivegreen}(44.10)}  &            31.89 $\pm$ 1.24 {\color{olivegreen}(33.20)}   \\
                                & Retrieval (Top1)      &            63.46 $\pm$ 2.30  {\color{olivegreen}(65.99)}  &            50.24 $\pm$ 2.21 {\color{olivegreen}(52.66)}   \\  \cmidrule{2-4}
                                & Llama-7B (Top3) ICL   &            65.61 $\pm$ 0.17  {\color{olivegreen}(65.80)}  &            36.50 $\pm$ 0.48 {\color{olivegreen}(37.01)}   \\
                                & ChatGPT (Top3) ICL    & \textbf{   68.91 $\pm$ 0.48  {\color{olivegreen}(69.38)}} & \textbf{   57.41 $\pm$ 0.40 {\color{olivegreen}(57.82)}}  \\  \midrule
            \multirow{4}{*}{2}  & BERT (Vanilla FT)     &            46.31 $\pm$ 0.65  {\color{olivegreen}(46.85)}  &            5.11  $\pm$ 1.31 {\color{olivegreen}(5.51)}    \\
                                & Llama-7B (Seq2Seq FT) &            45.66 $\pm$ 0.10  {\color{olivegreen}(45.77)}  &            41.26 $\pm$ 0.11 {\color{olivegreen}(41.39)}   \\
                                & Retrieval (Top1)      &            69.85 $\pm$ 0.63  {\color{olivegreen}(70.58)}  &            58.64 $\pm$ 0.58 {\color{olivegreen}(59.25)}   \\  \cmidrule{2-4}
                                & Llama-7B (Top3) ICL   &            67.09 $\pm$ 0.19  {\color{olivegreen}(67.29)}  &            54.70 $\pm$ 0.30 {\color{olivegreen}(55.03)}   \\
                                & ChatGPT (Top3) ICL    & \textbf{   71.68 $\pm$ 0.09  {\color{olivegreen}(71.76)}} & \textbf{   61.99 $\pm$ 0.10 {\color{olivegreen}(62.08)}}  \\  \midrule
            \multirow{4}{*}{4}  & BERT (Vanilla FT)     &            56.00 $\pm$ 4.25  {\color{olivegreen}(57.18)}  &            31.04 $\pm$ 16.65{\color{olivegreen}(33.77)}   \\
                                & Llama-7B (Seq2Seq FT) &            59.02 $\pm$ 0.08  {\color{olivegreen}(59.10)}  &            52.63 $\pm$ 0.07 {\color{olivegreen}(52.66)}   \\
                                & Retrieval (Top1)      &            75.37 $\pm$ 0.70  {\color{olivegreen}(76.08)}  &            65.94 $\pm$ 0.57 {\color{olivegreen}(66.41)}   \\  \cmidrule{2-4}
                                & Llama-7B (Top3) ICL   &            71.68 $\pm$ 0.09  {\color{olivegreen}(71.77)}  &            60.22 $\pm$ 0.04 {\color{olivegreen}(60.26)}   \\
                                & ChatGPT (Top3) ICL    & \textbf{   75.62 $\pm$ 0.15  {\color{olivegreen}(75.78)}} & \textbf{   66.34 $\pm$ 0.09 {\color{olivegreen}(66.41)}}  \\  \midrule
            \multirow{4}{*}{8}  & BERT (Vanilla FT)     &            66.24 $\pm$ 1.96  {\color{olivegreen}(67.53)}  &            50.21 $\pm$ 5.05 {\color{olivegreen}(52.60)}   \\
                                & Llama-7B (Seq2Seq FT) &            69.78 $\pm$ 0.05  {\color{olivegreen}(69.83)}  &            63.22 $\pm$ 0.04 {\color{olivegreen}(63.26)}   \\
                                & Retrieval (Top1)      & \textbf{   79.04 $\pm$ 0.48  {\color{olivegreen}(79.53)}} & \textbf{   70.59 $\pm$ 0.52 {\color{olivegreen}(71.04)}}  \\  \cmidrule{2-4}
                                & Llama-7B (Top3) ICL   &            75.03 $\pm$ 0.04 {\color{olivegreen}(75.07)}   &            63.58 $\pm$ 0.03 {\color{olivegreen}(63.61)}   \\
                                & ChatGPT (Top3) ICL    &            76.93 $\pm$ 0.05 {\color{olivegreen}(76.98)}   &            67.54 $\pm$ 0.04 {\color{olivegreen}(67.57)}   \\  \midrule
            \multirow{4}{*}{16} & BERT (Vanilla FT)     &            75.52 $\pm$ 0.32 {\color{olivegreen}(76.07)}   &            65.85 $\pm$ 1.28 {\color{olivegreen}(66.96)}   \\
                                & Llama-7B (Seq2Seq FT) &            78.42 $\pm$ 0.19 {\color{olivegreen}(78.66)}   &            70.09 $\pm$ 0.06 {\color{olivegreen}(70.15)}   \\
                                & Retrieval (Top1)      & \textbf{   81.12 $\pm$ 0.26 {\color{olivegreen}(81.38)}}  & \textbf{   73.72 $\pm$ 0.17 {\color{olivegreen}(73.82)}}  \\  \cmidrule{2-4}
                                & Llama-7B (Top3) ICL   &            76.43 $\pm$ 0.05 {\color{olivegreen}(76.48)}   &            65.56 $\pm$ 0.04 {\color{olivegreen}(65.60)}   \\
                                & ChatGPT (Top3) ICL    &            8.62  $\pm$ 0.03 {\color{olivegreen}(78.65)}   &            69.56 $\pm$ 0.03 {\color{olivegreen}(69.59)}   \\ 
        \bottomrule
    \end{tabular}
    }
    \caption{The Micro-F1 and the Macro-F1 scores of the Llama model on WOS dataset. We reported the average, standard deviation, and best results across three random seeds. \textbf{Bold}: best result.}
    \label{tab:llama}
    \end{table}

We also apply a powerful open access LLM base model Llama-7B for comparison.
The results are shown in Table \ref{tab:llama}.
\texttt{Llama-7B (Seq2Seq FT)} means we fine-tune the pre-trained Llama on our few-shot training set to generate hierarchical labels with a sequence-to-sequence target.
\texttt{Llama-7B (Top3) ICL} means we use the ICL with our retrieved top 3 demonstrations on the fixed Llama model without fine-tune.
The intricate architecture and extensive parameters of Llama-7B contribute to its superior performance over BERT (110M) in fine-tuning scenarios.
In contrast, our retrieval model, built upon the BERT architecture with 110M parameters, consistently outperforms the fine-tuned results of Llama-7B.

In extremely few shot settings (such as Q$=1, 2, 4$), applying ICL on LLM with our retrieved results leads to further performance improvement, with more powerful models like ChatGPT typically demonstrating superior results.
When Q grows, our retrieval methods could outperform LLM-based ICL.
\texttt{Llama-7B (Top3) ICL} shows only marginal improvement compared to the \texttt{Retrieval (Top1)} result in 1-shot setting,
implying that the degree of enhancement in ICL inference results is contingent upon the performance strength of the LLM.

\begin{figure}[t]
    \centering
    \includegraphics[width=0.90\columnwidth]{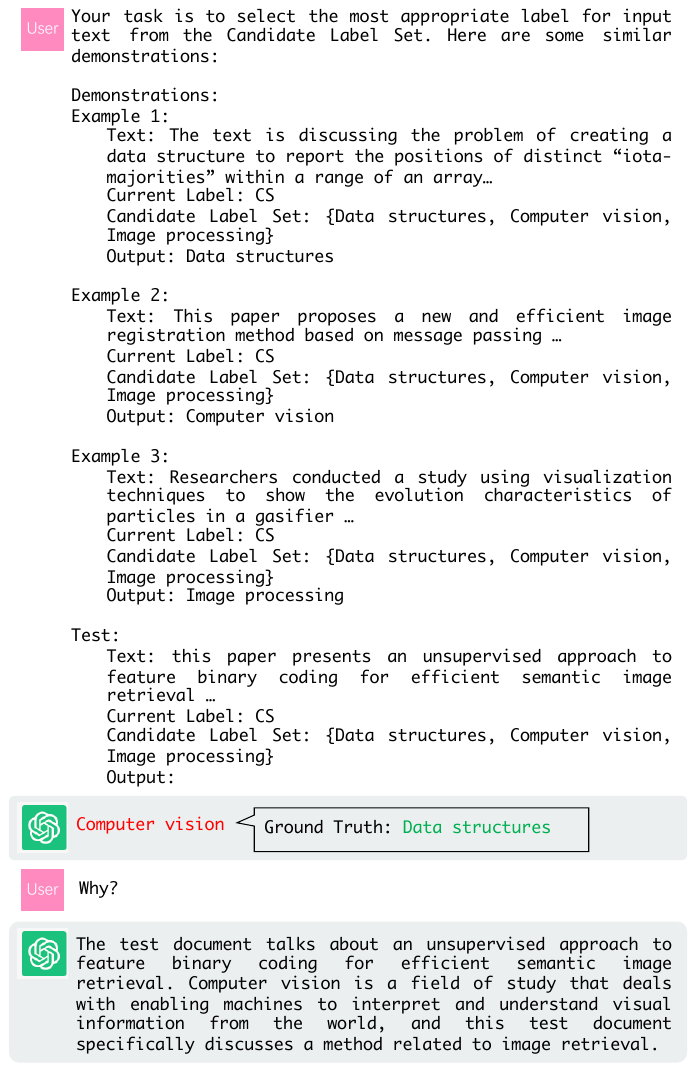}
    \caption{A case that LLM fails to choose the correct label although the retrieved Top1 result is right.}
    \label{fig:case}
\end{figure}

\paragraph{The improvement limitation of ICL inference.}
We present the Top1 and Top3 retrieval results\footnote{Top3 selects the label with the highest overlap with the gold-standard label among the top3 retrieved labels as the predicted label result.} in Figure \ref{fig:ICL}.
For the retrieval-style ICL method, if we only provide the Top1 example retrieved,
the ICL inference result will be consistent with Top1 retrieval result.
Therefore, we present the results of \texttt{Retrieval Top2 + ICL} and \texttt{Retrieval Top3 + ICL} in Figure \ref{fig:ICL}.

We show the results of constructing the candidate set without employing the filtering strategy, labeled as \texttt{Retrieval Top3 w/o filter + ICL}.
When $Q=1$, the Top3 retrieved labels are unique, rendering the results identical to \texttt{Retrieval Top3 + ICL}.
When $Q=2$, the Top3 retrieved labels typically encompass only two categories, leading to a scenario where one label (often the Top1 label) appears twice in the demonstration selections, introducing a bias in the inference process.
As a result, \texttt{Retrieval Top3 w/o filter + ICL} slightly underperforms compared to \texttt{Retrieval Top2 + ICL}.
When $Q\geq4$, the Top3 retrieved labels usually belong to a single category, aligning the outcomes with those of \texttt{Retrieval Top1}.

Ideally, the ICL method can select the label closest to the gold-standard label from the candidate label set based on the provided examples.
Taking Top3 as an example, the \texttt{Retrieval Top3 + ICL} curve should be close to the \texttt{Retrieval Top3} curve.
In fact, the result curves of ICL are all round \texttt{Retrieval Top1},
and the curve of \texttt{Retrieval Top2 + ICL} is closer to \texttt{Retrieval Top1} than \texttt{Retrieval Top3 + ICL}.
We analyze several possible reasons as follows:
(1) Firstly, the LLM is not fine-tuned, and its understanding of labels may be inconsistent with the training set.
(2) The effect of ICL is limited. The quality of retrieval examples is getting strong, resulting in increasingly similar candidate labels provided, which may increase the difficulty of the LLM inference.
Figure \ref{fig:case} shows a case that LLM fails to choose the correct label, although the retrieved Top1 result is right.
(3) Although LLM has demonstrated strong ability, there is still room for improvement. Using a more powerful LLM may yield better results.
It can be concluded that our retrieval-style ICL method is far superior to direct inference using LLM,
and can improve the performance on retrieval-based inference under extremely low resources.
However, enhancing the retrieval results cannot continuously improve the performance of ICL.

\begin{table}[t]
    \centering
    \resizebox{1\columnwidth}{!}{
        \begin{tabular}{cl cc }
            \toprule



            \multirow{2}{*}{\textbf{Q}}          &                          & \multicolumn{2}{c}{\textbf{WOS}}  \\ \cmidrule{3-4}
                                                 &                          &     Micro-F1    &     Macro-F1    \\ \midrule

            \multirow{2}{*}{0}  & LLM + iterative          &\textbf{55.40}     &\textbf{44.79}   \\
                                & -w/o iterative           &        26.70      &        16.44    \\ \midrule

            \multirow{6}{*}{1}  & Random Samples + ICL     &        56.42      &        45.24    \\
                                & Retrieval (Top3) + ICL   &\textbf{68.91}    &\textbf{57.41}   \\
                                & -w/o iterative           &        68.52      &        57.06    \\
                                & -w/o similar samples     &        64.75      &        52.27    \\
                                & -w/o pruning             &        60.37      &        47.43    \\
                                & -w/o candidate label set &        52.35      &        32.66    \\

        \bottomrule
    \end{tabular}
    }
    \caption{Results of different prompt settings on WOS. \texttt{w/o} means 'without'. \textbf{Bold}: best result.}
    \label{tab:prompt}
    \end{table}

\paragraph{The impact of different prompts on LLM inference.}
The differences in prompts directly affect the results of LLM inference.
We conduct ablation experiments on the prompts we proposed to verify the rationality of our iterative prompts,
and the results are shown in Table \ref{tab:prompt}.

Under the zero-shot setting, when compared to directly inputting all hierarchical label paths to the language model, the iterative method improves the Micro-F1 by 28.70\% and the Macro-F1 by 28.35\%.
It helps alleviate the negative impact of excessively long prompts during the inference process.
This indicates that the iterative method is particularly effective for handling HTC tasks.

Taking 1-shot as an example, we randomly select three samples in training dataset to form the prompt and use all labels from the target hierarchy layer as the candidate label set for iterative prediction.
Interestingly, even with prompts constructed from random samples, the results obtained through ICL outperform \texttt{LLM + iterative}.
This finding emphasizes the effectiveness of the ICL approach in generating inference results that closely match the desired format.

Then, we use our Top3 retrieval reulst as demonstrations and conduct four ablation comparisons.
The first one is to remove the iterative operation,
which means that the candidate label set consists of label paths,
and all hierarchical labels are predicted at once.
The second one is to remove all similar samples,
and only provide the current label and candidate label set of the test document.
The third one is to remove the pruning operation,
which means that the candidate label set consists of all child labels of the current label.
The last one is to remove the candidate label set and let the LLM select the most similar text,
and use the label of the similar test as the label of the test document.
The results prove that our prompts is reasonable,
and each part of the prompt has a positive effect on inference\footnote{All details of prompts will be publicly available, thus enhancing the reproducibility of our work.}.

\paragraph{Retrieval-assisted human annotation.}
Furthermore, we recruit non-experts to annotate a portion of the test dataset, aiming to ascertain the expected upper-bound performance.
We conduct experimental analyses on the WOS dataset as examples.
We recruit college students with proficient English and conduct a simple annotation test,
selecting nine annotators with comparable levels of annotation skill and efficiency,
who are then divided into groups of three for subsequent annotation tasks.
We randomly select 200 instances from the WOS test dataset for annotation.
Three annotation methods are employed:
(1) providing only the full list of labels;
(2) based on (1), supplying an explanation for each label;
(3) based on (2), offering the Top3 similar examples assisted by a retriever model trained on the 16-shot setting for annotation.
Each annotation method is carried out by three annotators, with the final annotation results produced by voting.

\setlength{\tabcolsep}{5pt}
\begin{table}[t]
\begin{center}
\resizebox{1\columnwidth}{!}{
\begin{tabular}{lcc}
\toprule
\multirow{2}{*}{\textbf{Annotation Methods}} & \multirow{2}{*}{\textbf{Micro-F1}} & \multirow{2}{*}{\textbf{Avg. Time (s)}} \\
& &  \\
\midrule
(1) direct classification        & 67.25  & 37.2    \\
(2) with label description       & 73.00  & 45.8    \\
(3) 16-shot retrieval-assisted   & 86.44  & 9.1     \\

\bottomrule
\end{tabular}
}
\caption{Statistical results of different annotation methods on WOS. Avg. Time indicates the average time (s) spent annotating each instance.}
\label{tab:human_annotated}
\end{center}
\end{table}

The statistical results of different annotation methods are presented in Table \ref{tab:human_annotated}.
Annotation method (1) represents the upper-bound result based on human knowledge under the 0-shot setting.
A comparison reveals that after providing label descriptions, the Micro-F1 increases by 5.75\%,
but the average annotation time also lengthens due to the provision of more information.
When assisted by the Top3 examples provided by a retriever trained in the 16-shot setting, the Micro-F1 significantly improves by 13.44\%, and the average annotation time is reduced to only one-fifth of that for method (2), as many clearly incorrect labels are eliminated, reducing the difficulty of annotation.
This indicates that our retrieval method can assist human annotation, effectively improving the quality of human annotations and reducing the time required.

\begin{figure}[t]
    \centering
    \includegraphics[width=0.90\linewidth]{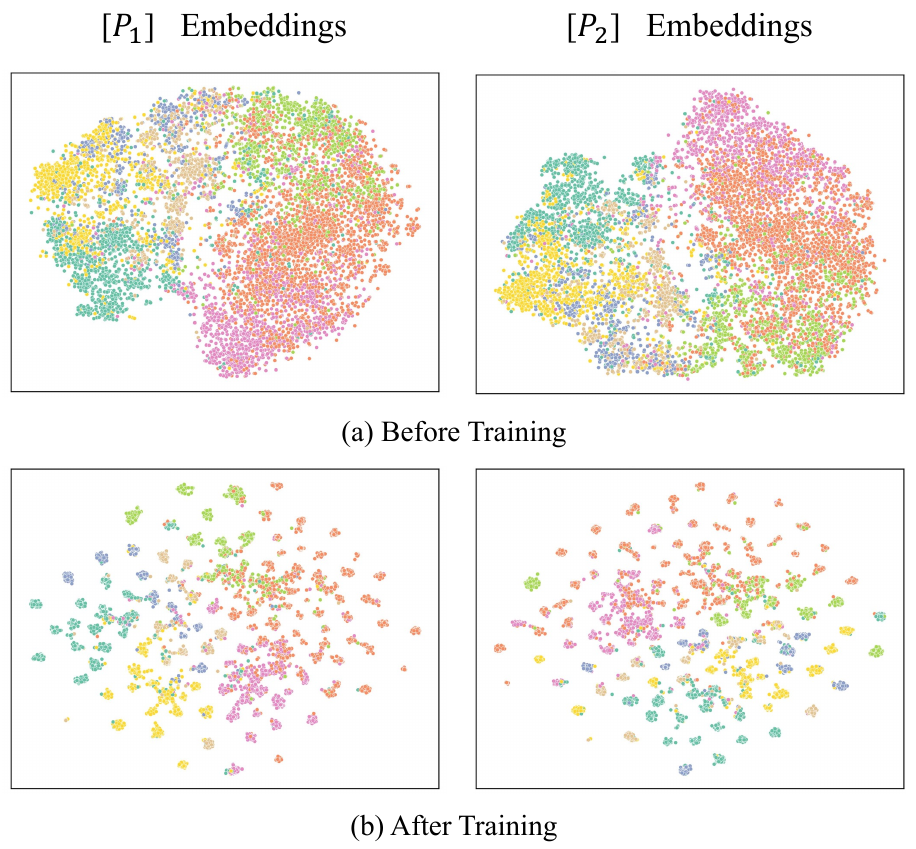}
    \caption{Visualization on the WOS test dataset. The top two figures show \texttt{[P]} embeddings obtained using the original BERT, while the bottom two figures show \texttt{[P]} embeddings obtain after training by our method.}
    \label{fig:visual}
\end{figure}

\paragraph{Visualization of index vector.}

Finally, we use T-SNE \cite{van2008visualizing} to visualize the changes in \texttt{[P]} of the WOS test dataset before and after training, as shown in Figure \ref{fig:visual}.
We find that index vectors exhibit clear hierarchical clustering characteristics, further demonstrating the effectiveness of our method.

\section{Conclusion}
In this paper, we proposed a retrieval-style ICL framework for few-shot HTC.
We uniquely identify the most relevant demonstrations from a retrieval database to improve ICL performance and meanwhile designed an iterative policy to inference hierarchical labels sequentially, significantly reducing the number of candidate labels.
The retrieval database is achieved by using a HTC label-aware representation for any given input, enabling the differentiation of semantically-closed labels (especially the leaf adjacent labels).
The representation learning is implemented by continual training on a PLM with three carefully-designed objectives including MLM, layer-wise classification, and a novel DCL objective.

We conducted experiments on three benchmark datasets to evaluate our method.
The results show that our method is highly effective, which is able to gain large improvements among a serious of baselines.
Finally, our method can bring the state-of-the-art results in few-shot HTC on the three datasets.
Further, we performed comprehensive analysis for deep understanding of our method,
spreading various important factors.

This work still includes several unresolved problems, which might be addressed in the future.
Firstly, LLMs are currently confined to expanding text via label descriptions and their application to full training set expansion has not been effective.
In order to fully utilize LLMs in text expansion, we need further optimization.
Second, the performance gap between supervised methods and our ICL-based approach appears to diminish with increasing training dataset size, suggesting the need for further analysis.

\section*{Acknowledgements}
We thank the anonymous reviewers for their helpful comments. This work is supported by the National Natural Science Foundation of China (No. 62176180).

\bibliography{tacl2021}
\bibliographystyle{acl_natbib}

\end{document}